\documentclass[english]{Liebert_Author}

\usepackage{amsmath}
\usepackage{amssymb}
\usepackage[T1]{fontenc}
\usepackage[utf8]{inputenc}
\usepackage{babel}
\usepackage{geometry}
\usepackage{algorithm}
\usepackage{algorithmic}
\usepackage{xcolor}
\usepackage{soul}
\newcommand{\newtext}[1]{{\color{black} #1}}

\usepackage{booktabs}

\usepackage[nameinlink,capitalize]{cleveref}
\crefname{section}{Sec.}{Secs.}
\crefname{table}{Table}{Tables}

\title{Parallel Simulation of Contact and Actuation \\for Soft Growing Robots}

\author{Yitian Gao${}^{1*}$, Lucas Chen${}^{1*}$, Priyanka Bhovad${}^{2}$, Sicheng Wang${}^{2}$,\\
Zachary Kingston${}^{1}$, and Laura H. Blumenschein${}^{2\dagger}$\\
{${}^{1}$ Department of Computer Science, Purdue University}\\
{${}^{2}$ Department of Mechanical Engineering, Purdue University}\\
{${}^{*}$ Equal Contribution}\\
{${}^{\dagger}$ To whom correspondence should be addressed;}\\
{E-mail: \url{lhblumen@purdue.edu}}
}

\begin{document}
\maketitle


\newpage

\begin{abstract}
Soft growing robots, commonly referred to as vine robots, have demonstrated remarkable ability to interact safely and robustly with unstructured and dynamic environments. 
It is therefore natural to exploit contact with the environment for planning and design optimization tasks. 
Previous research has focused on planning under contact for passively deforming robots with pre-formed bends. 
However, adding active steering to these soft growing robots is necessary for successful navigation in more complex environments.  
To this end, we develop a unified modeling framework that integrates vine robot growth, bending, actuation, and obstacle contact. 
We extend the beam moment model to include the effects of actuation on kinematics under growth and then use these models to develop a fast parallel simulation framework. 
We validate our model and simulator with real robot experiments. 
To showcase the capabilities of our framework, we apply our model in a design optimization task to find designs for vine robots navigating through cluttered environments, identifying designs that minimize the number of required actuators by exploiting environmental contacts.
We show the robustness of the designs to environmental and manufacturing uncertainties. 
Finally, we fabricate an optimized design and successfully deploy it in an obstacle-rich environment.
\end{abstract}

\section{Introduction}

Soft growing robots, also known as vine robots, are a class of continuum robots that extend from the tip by pressure-driven eversion~\cite{hawkes2017soft}. 
Since growth isolates the robot from the environment, vine robots have shown significant beneficial behaviors in tasks with natural pathways to follow, such as medical procedures~\cite{girerd2024material,borvorntanajanya2024model,berthet2021mammobot}, \newtext{especially for colonoscopies and endoscopies~\cite{colonoscope, endoscope, InchIGRAB}}, pipe inspection~\cite{qin2024design}, and archaeology~\cite{coad2019vine}. 
These behaviors arise from the passive buckling of thin inflated tubes~\cite{comer_levy,haggerty19}, a challenging feature to accurately model.

Multiple works have examined \newtext{kinematic models to predict} and use this passive deformation~\cite{greer20, fuentes23}. 
However, \newtext{the heuristic models in} these works have been limited to purely passive behavior, primarily due to the increased difficulty of modeling the effects of active steering.
Other works have also addressed the general simulation of vine robot growth with \newtext{at most} some pre-formed deformations, including fast, kinematic-only models~\cite{fuentes25} and dynamic simulations~\cite{jitosho2021dynamics,li2021bioinspired}.
While slow and accurate FEM simulators~\cite{faure2012sofa} have been used to model actuated vine robots~\cite{du2023finite}, these are inefficient for downstream tasks such as planning, control, and design that require high-throughput simulation.
Recent work has addressed fast parallel simulation of these robots that accurately captures the bending and buckling behavior intrinsic to vine robots~\cite{robosoft_diff_sim}.
However, no simulation framework has addressed efficient modeling of actuated vine robots under contact forces.

\newtext{
While previous studies have captured the obstacle interactions of vine robots under passive deformation, this work achieves the integration of all key capabilities, including growth, pneumatic actuation, beam mechanics, and environmental contact. In this work, we provide for the first time the tools to design, model, control, and plan actuation for vine robots in contact-rich, complex environments. To validate our framework, we deploy the optimized designs in real environments and compare the performance with simulated predictions. Our major contributions include:

\begin{itemize}
\item \textbf{Unified modeling of growth, steering actuation, bending, and contact}:
We present a new modeling framework that captures how vine robots buckle, bend, deform, and extend while growing into obstacles with active steering. We choose series pneumatic artificial muscles (sPAMs)~\cite{greer2017series} as bending actuators for this study. Our modeling framework independently considers the effects of the actuator and inflated body when calculating the moment balance discretized along the length of the vine robot. Importantly, this leverages a continuous form of the thin-walled inflated tube buckling model that includes partial buckling reactions under small angle deflection~\cite{wang2024anisotropic} called the wrinkling-based bending moment model. Thus, we can predict both distributed actuation seen during steering and concentrated actuation seen under environmental contact. \cref{sec:model} combines the models of sPAM actuation, wrinkling-based restoring moment to calculate the net restoring moment that the vine robot segment experiences.

\item \textbf{Fast parallel simulation framework}:
GPU-parallel simulation has been enabled by efficient parallel computation libraries such as PyTorch and JAX. We build upon a prior parallel simulation framework~\cite{robosoft_diff_sim} for fast and accurate prediction of actuated vine robots navigating cluttered environments. We use our novel combined actuation, bending, and contact model, and switch to a faster gradient-based penalty method to run a batch of simulations in parallel with orders of magnitude speedups. \cref{sec:sim} describes the details of position-based dynamic simulation. To perform fast computations of sPAM actuators, a neural surrogate model is developed as detailed in \cref{surrogate}. \cref{design_optimization} describes the design optimization framework and the planner algorithm for batched SST$^*$.

\item \textbf{Application for long-horizon design optimization task}:
To demonstrate the capability of our model for long-horizon planning, we apply it in a design optimization task: finding vine robot designs that minimize the number of actuators needed to navigate a complex, cluttered environment. The resulting designs successfully reach the target location when deployed in the real environment, demonstrating real-world applications of our framework. \cref{sec:exp_res} describes the experimental and simulation results while analyzing the funneling and robustness behaviors.
\end{itemize}

In summary, our work presents a unified computational framework to optimize the design of vine robots. These advances enable the implementation of simulation tools to deploy optimized designs in real, cluttered environments. We release our simulator and design optimization tool as open source: \url{https://github.com/CoMMALab/ActVineSimPy}.}

\section{Related Work}

The flexible, yet inextensible material used in vine robots necessitates the use of novel approaches for 
\newtext{{\textbf{modeling and simulation of vine robot}}} behavior. 
While early modeling focused primarily on describing the growth~\cite{blumenschein2017modeling}, first-principles models for vine robot buckling have been built on inflated thin-shelled beam models~\cite{comer_levy,fichter66}.
Discrete buckling models, which predict restoring moment independent of bend angle~\cite{haggerty19,hwee23}, led to kinematic models for contact-induced bending~\cite{greer20}, tip localization~\cite{frias_miranda23}, and obstacle mapping~\cite{fuentes23}. Recent work has further strengthened this model by considering partial buckling~\cite{wang2024anisotropic}.

For distributed surface wrinkling, these constant moment models are not as accurate, so heuristic models have been developed instead, using experimental parameter fitting to create kinematic models~\cite{greer19}, or geometrically relating surface strain to resulting general curves~\cite{blumenschein22,wang22}.
The geometric mappings are generally limited to quasi-static behaviors, though data-driven approaches with Koopman operators have shown that dynamic behaviors can also be empirically modeled~\cite{haggerty2023control}.
While combined system models have been primarily heuristic, some first-principles actuator models have been developed for series Pneumatic Artificial Muscles (sPAMs)~\cite{greer19} and a range of other common and vine robot-specific actuators~\cite{kubler19}. 

While a\newtext{c}tuator-beam models describe free-space actuated vine robots, incorporating contact with the environment is challenging.
Finite element models built on frameworks like SOFA~\cite{faure2012sofa} have been used to successfully model vine robots~\cite{li2021bioinspired,du2023finite,vartholomeos2024lumped,wu2023towards}.
However, these approaches are computationally expensive and ill-suited for real-time, at-scale simulation.
Other work has simplified the simulations to rigid body models, either using minimal coordinates through virtual joint angles~\cite{el2018development} or maximal coordinates with poses of frames in a global reference with implicit constraints~\cite{jitosho2021dynamics}.
Previous work has proposed more realistic models for computing strain that respect geometric design parameters~\cite{wang22,sPAM} and has made these scalable using GPU-accelerated computation frameworks~\cite{robosoft_diff_sim}.
Overall, our work fills a gap for efficient models that generalize to external and internal forces. 

Our 
\newtext{\textbf{design optimization}} approach is built upon sampling-based kinodynamic planning (SBKP)~\cite{lavalle2006planning,donald1993,li2016}, which addresses the general problem of finding a sequence of controls that, when applied to a robot, achieves a goal state.
Unlike previous works that use planning to design a vine robot (e.g.,~\cite{selvaggio20}), we are given only black-box forward dynamic simulation that captures the vine robot evolution given actuation and environment interactions. 
Some planners, such as Kinodynamic-RRT* for linear dynamics~\cite{webb2013}, arbitrary kinodynamic planning in combination with the AO-X meta algorithm~\cite{Hauser2016} and Stable Sparse RRT (SST)~\cite{li2016}, provide asymptotic optimality guarantees, but with no guarantee on when high-quality solutions will be found.
Thus, in practice, accelerating the speed of dynamic simulation, the bottleneck of SBKPs, helps them find better solutions faster.
Recent work such as Kino-PAX~\cite{10844318} has focused on highly parallel implementations to address these computational challenges.
Our approach also leverages the large-scale batch GPU computation of our proposed simulator to improve performance.

\section{Modeling of Combined Actuation and Growth}
\label{sec:model}

Accurately modeling actuated soft growing robots under environmental contact presents a challenge due to the non-linear coupling between actuator force and displacement, beam stiffness or restoring moment, and contact forces.
While an actuated robot deployed in free space can be modeled heuristically as one or more constant curvature segments with radius of curvature inversely proportional to actuator pressure~\cite{greer19}, and the path of an unactuated vine robot under environmental contact can be robustly predicted based on the visibility graph of the obstacles~\cite{fuentes25}, these heuristic models do not obviously or easily combine.
To address this complexity, we develop a unified modeling approach which builds on analytical models of each component separately and then discretizes the vine robot along its length to locally capture the effects of contact and actuation. 
With this model, the vine robot's dynamics can be simulated in cluttered environments, where contact interactions significantly influence robot behavior and thus can be exploited for improved navigation performance. 
In this section, we present each component model separately and then discuss the method to combine the models to accurately predict both actuation and contact with the environment.

\subsection{Serial Pneumatic Artificial Muscles (sPAMs)}
\label{sPAM_theory}

We employ serial Pneumatic Artificial Muscles (sPAMs) to apply distributed actuation along the robot body.
A sPAM can be fabricated using a pliable thin-film tube, which has an inflated radius $R_\text{act}$, and constricting that tube to radius $R_c$ (where $R_c \ll R_\text{act}$) at regular intervals $l_0$, forming a chain of bubbles.
When the actuators are completely deflated, they start at full extension (\cref{fig:variables}), with constrictions at interval lengths of $l_0$. Upon pressurization, the actuator expands radially and, as a result, contracts. The radial expansion continues until the maximum radius at the actuator unit reaches the limit, $R_\text{act}$, saturating the response for that part of the actuator cell. Further pressurization increases this saturated region, further shortening the actuator and decreasing the active region. 

To make an actuator component model, we build upon the refined sPAM model in~\cite{sPAM}, which adapts the ideal PPAM (Pleated Pneumatic Artificial Muscle) model in~\cite{daerden99PPAM}. 
The ideal PPAM model does not account for the saturation based on the tube radius, only predicting the cross-section profile which maintains the material length and maximizes volume.
The improved sPAM model incorporates actuator saturation by moving the saturated length to shrink the effective length of the actuator when predicting the force-strain output. For a given set of actuator design parameters and pressure inputs:
\begin{subequations}\label{eq:sPAM_model}
\begin{align}
&\frac{E(\phi_{Rc},m)-\frac{1}{2}F(\phi_{Rc},m)}{\sqrt{m}\cos\phi_{Rc}} = \frac{l_a}{2R_c}\left(1 -\frac{l_0}{l_a}\varepsilon\right), \label{eq:sPAM_1}\\
&\frac{F(\phi_{Rc},m)}{\sqrt{m}\cos\phi_{Rc}}=\frac{l_a}{R_c}\left(1+\frac{a}{2m\cos^2\phi_{Rc}}\right), \label{eq:sPAM_2}\\
&a = \frac{P_\text{act}R_c}{2tE} \,,\quad \phi_{Rc} = \cos^{-1}\left(\frac{R_c}{R_\text{act}}\right) \label{eq:phiRC}\\
&F_t = \pi{P_\text{act}}R_{c}^2\frac{1-2m}{2m\cos^2\phi_{Rc}}. \label{eq:sPAM_force}
\end{align}
\end{subequations}
Here, $\varepsilon$ denotes the strain, $F_t$ represents the contraction force.
$m\in(0,0.5]$ and $\phi_{Rc}\in(0,\frac{\pi}{2}]$ are dimensionless model parameters, with the former denoting the extent of contraction relative to the full contraction and the latter proportional to the maximum radius of the sPAM unit at a given state.
$l_a$ and $l_0$ denote the active and total actuator lengths respectively, and $F(a,b)$ and $E(a,b)$ are incomplete elliptic integrals of the first and second kind.
The pressure-dependent correction factor $a$ relates the actuator's material properties to force-strain computation.
While in the original PPAM model this correlated with material elasticity, $E$, and pressure, $P_\text{act}$, with the relatively inextensible tube and constraint materials we instead treat this parameter as an empirically determined constant ($a=0.0001$).
The model accounts for saturation effects by distinguishing between $l_a$ and $l_0$ and by setting $\phi_{Rc}$ based on the saturation and constraint radii (\cref{eq:phiRC}). 
Setting $l_a = l_0$ and treating $\phi_{Rc}$ as unknown instead recovers the PPAM model applicable to unsaturated scenarios.
To generate the force-strain output pairs needed for the neural surrogate modeling described in~\cref{surrogate}, we vary $m$ over $m \in (0, 0.5]$. 
For each $m$, we solve the governing equation for $\varepsilon$, and in the unsaturated case, for $\phi_{Rc}$. 
We then compute the contraction force $F_t$ using these solutions and the prescribed design parameters.

\subsection{Wrinkling-based Restoring Moment Model}
\begin{figure}[H]
    \centering
    \includegraphics[width=0.78\linewidth]{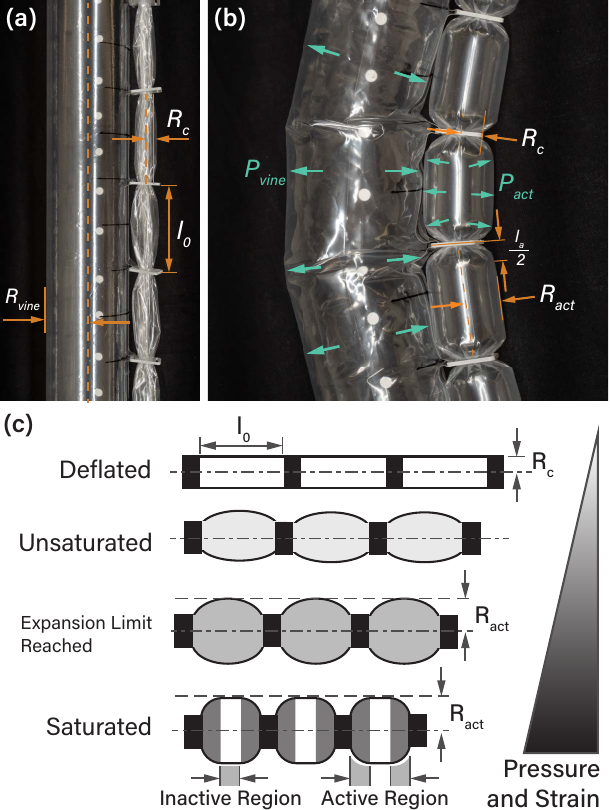}
    \caption{\textbf{(a), (b)}: Variable definitions for the sPAM and inflated beam model.
    \textbf{(c)}: Illustration of sPAM states from completely deflated with zero strain (top) to saturated with maximal strain (bottom).
    Darker color indicates higher input pressure and greater contraction.}
    \label{fig:variables}
\end{figure}

\begin{figure}[H]
    \centering
    \includegraphics[width=\linewidth]{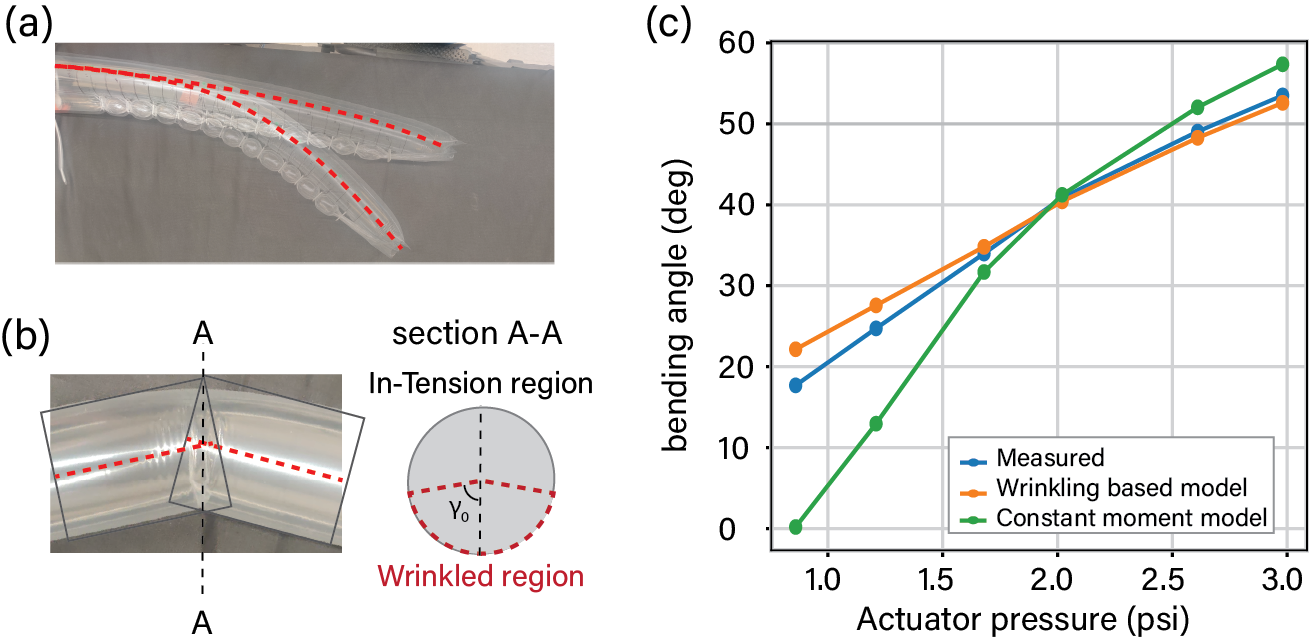}
    \caption{\textbf{(a)}: Vine segment bending under sPAM actuation.
    \textbf{(b)}: Vine segment wrinkled and tensioned regions depicting $\gamma_0$.
    \textbf{(c)}: Comparison of measured bending angles with theoretical estimations using wrinkling based model and constant moment model.}
    \label{fig:wrinkling_model}
\end{figure}

The body of the vine robot is made of a pressurized thin-film tube, which we can treat as an inflated beam. 
An inflated beam model developed by Comer and Levy~\cite{comer_levy} demonstrates that the restoring moment of a bent inflated beam increases with surface wrinkling development at bending locations.
Previous implementations of this model have focused on the \textit{maximum} restoring moment when the surface is fully wrinkled, especially in works which discuss the buckling of the vine robot under environmental forces~\cite{greer20}. 
However, for small deflection angles ($\lesssim 10$ degrees), like those seen in steering (\cref{fig:wrinkling_model}a), this significantly overestimates the restoring moment.
A wrinkling-based model described in~\cite{robosoft_diff_sim} captures this angle-to-moment relationship by introducing a critical surface strain $\varepsilon_\text{critical}$ at which wrinkling initiates. The critical strain is obtained by fitting experimentally measured deflection-restoring moment data.
The restoring moment $M$ is computed as:
\begin{subequations}\label{wrinkling_model}
\begin{align}
&M = \pi P_\text{vine}R_\text{vine}^3 \frac{\sin(2\gamma_0) + 2 \pi -2\gamma_0}{4 (\sin\gamma_0+(\pi - \gamma_0)\cos\gamma_0)},\\
&\gamma_0 = \cos^{-1}\left(2\frac{\varepsilon_\text{critical}}{\sin(\theta_\text{curv}/2)} - 1\right). \label{eq:gamma}
\end{align}
\end{subequations}
Here, $P_\text{vine}$ denotes the pressure of the vine robot and $R_\text{vine}$ denotes the radius of the vine robot as shown in~\cref{fig:variables}. 
$\theta_\text{curv}$ denotes the turning angle at the point of wrinkling. 
$\gamma_{0}$ represents the angle corresponding to the wrinkled surface of the vine robot as shown in~\cref{fig:wrinkling_model}b and defined in~\cref{eq:gamma}.
Importantly, this model calculates the restoring moment as proportional to the total axial material tension, summed around the tube circumference, i.e. $\pi P_\text{vine} R_\text{vine}^2$. If the vine robot is growing, $P_\text{vine}$ cannot exceed the conditions of quasi-static equilibrium for growth, which are based on the vine robot material and geometry, the internal material tension, and the growth speed~\cite{blumenschein2017modeling}. In practice, the effects of growth speed are negligible, so a value for $P_\text{vine}$ can be set by modifying the internal material tension.

\subsection{Combined Actuation and Contact Model}
\label{combined_model}

To combine the component models, we can recognize that the moments exerted by the sPAMs and by the bending of the vine robot segment should balance in free space, and combine to produce a net restoring moment under dynamic conditions.
Without external contact, this equilibrium relationship enables determination of both the strain produced in the vine segment and the radius of curvature in free space.
Compared with previous approaches, the wrinkling-based model improves the accuracy of curvature determination, as shown in~\cref{fig:wrinkling_model}c.

For developing the forward dynamics simulation that includes contact with the environment, the vine robot is discretized into rigid, serially connected segments. %
\newtext{This allows the bending behavior of each segment to be considered separately. The segments have a length of  $l_\text{seg} = 25$mm, excluding the last growing segment $l_\text{last}$, which grows with a predefined growth rate. The value 25mm was chosen for the segment length because it afforded a good balance between granularity and computational cost, allowing enough discretized segments to provide a reasonably accurate model without introducing too much extra computation.}

If the relative bending angle between any two consecutive segments of the vine robot is $\theta$, then the actuator strain $\varepsilon$ corresponding to the bending angle $\theta$ and net restoring moment $M_\text{tot}(\theta)$ for that segment at each time step is calculated as follows:
\begin{subequations}\label{eq:restoring_moment}
\begin{align}
&\varepsilon = \frac{(2R_\text{vine} + R_\text{act}){\theta}}{l_\text{seg}}, \\
& M_\text{tot}({\theta}) = M(\theta) - F_t(m,\phi_{Rc},P_\text{act})\cdot(2R_\text{vine}+R_\text{act}).
\end{align}
\end{subequations}
Here, the restoring moment of the vine robot segment $M(\theta)$ is calculated using~\cref{wrinkling_model}. 
The actuator force $F_t$ and the geometric parameters $(m, \phi_{Rc})$ for the actuator are calculated using~\cref{eq:sPAM_model}.

\section{Simulation of Actuation and Growth}
\label{sec:sim}

Although the individual bending, growth, and contact behaviors are well defined, producing a realistic simulation requires unifying these forces into a single forward function. 
We combine these forces in a cost-minimization problem, which we use to solve for the next state with a Position-Based Dynamics (PBD)~\cite{bender2014survey} simulation method, which has shown success for rigid and soft body simulations alike~\cite{xpbd}.
The pipeline leverages GPU-accelerated batch processing to achieve the throughput requirements necessary for effective design optimization in complex environments. \newtext{\cref{alg:sim_step} details the single simulation step that updates the vine robot position forward in time. This step essentially combines the effects of vine robot growth, bending, actuation, and obstacle contact to generate the next state of the vine robot.}

\subsection{Position-Based Dynamic Simulation}

As described in~\cref{combined_model}, we chose to discretize the vine into a chain of bodies to produce a finite parameterization. 
We aim to find the dynamics $\{\theta\}_{t+1}, l_\text{last} = f(\{\theta\}_t, l_\text{last})$. 
$l_\text{last}$ is the length of the distal segment; all others are held at a fixed length $l_{seg}$.
$\{\theta\}$ is the set of relative bending angles between consecutive bodies in the vine; they describe the local curvature at each point. 
This rigid body representation was also used in~\cite{jitosho2021dynamics, robosoft_diff_sim}, albeit with maximal coordinates.
Note that maximal coordinates require extra constraints to ensure that successive bodies remain connected; our minimal coordinate formulation can only produce connected vines. 
Here, we formulate the simulation as finding the next state with minimum cost:
\begin{equation}\label{pbd}
\text{sim}(\{\theta\}^*) = 
    \arg\min_{\{\theta\}} w_a M_\text{tot}(\{\theta\})
     + w_b\, \text{growth}(l_\text{last})
     + w_c\, \text{contact}(\{\theta\}, l_\text{last}, \mathcal{X}_{\text{obs}}),
\end{equation}
where ``\text{growth}'' tends to zero as $l_\text{last}$ increases with a predefined growth rate, ``\text{contact}'' measures the amount of penetration of bodies into obstacles, and $M_\text{tot}$ is the restoring moment of the current state as found in~\cref{eq:restoring_moment}. 
Often, such forces will be in opposition---for example, when the vine is growing into an obstacle, then the growth term will attempt to lengthen $l_\text{last}$ while ``\text{contact}'' will attempt to shorten it.
We empirically scaled the weights $w_a, w_b,$ and $w_c$ to ensure that in all common cases, contacts dominate over moments which in turn dominate growth, thus forbidding segments from growing into the obstacles, with $w_a = 160$, $w_b = 15$, $w_c = 1$. \newtext{An exception to this occurs when the tip gets stuck in a corner or head-on collision, causing growth to dominate bending moments as the vine coils up on itself.}
Note that we omit velocity and inertial terms as they are insignificant due to the high air resistance and low mass of the vine.

\cref{pbd} is highly nonlinear and the relative magnitudes of each force can change rapidly. 
We use %
\newtext{a penalty-based gradient method which iteratively searches for the minimization of \cref{pbd}}. %
We compute gradients using JAX autodifferentiation~\cite{bradbury2018jax} \newtext{and found 50 steps to be sufficient for convergence in all environments. The details of the implementation are presented in \cref{alg:sim_step}}.

\begin{algorithm}
\caption{\newtext{Single Simulation Step}}
\label{alg:sim_step}
\begin{algorithmic}[1]

\STATE \textbf{Input:} Initial state $\{\theta\}_t$, last length $l_{\text{last}}$, cost weights $w_a$, $w_b$, $w_c$ \\ 
\quad\quad neural surrogate parameters $\epsilon$, $l_0$, learning rate $\alpha$, obstacle set $\mathcal{X}_{\text{obs}}$

\STATE Initial vine state: $\{\theta\} \leftarrow \{\theta\}_t$
\STATE $m, \phi_{rc} \leftarrow \text{neural surrogate}(\epsilon, l_0)$
\STATE Actuator force $F_t \leftarrow ~\cref{eq:sPAM_force} (m, \phi_{rc})$

\STATE \quad

\FOR{$k = 1$ to $50$}
    \STATE Restoring moment $M \leftarrow ~\cref{wrinkling_model}(\{\theta\})$ 
    \STATE Total moment $M_{\text{tot}} \leftarrow ~\cref{eq:restoring_moment}(M, F_t)$
    \STATE Optimization Objective: 
    \[
        J = w_a M_{\text{tot}} + w_b \text{growth}(l_{\text{last}}) + w_c \text{contact}(\{\theta\}, l_{\text{last}}, \mathcal{X}_{\text{obs}})
    \]
    
    \STATE Gradient: $\nabla_{\theta} J \leftarrow$ autodifferentiate($J, \theta$)

    \STATE Gradient descent step: $\{\theta\} \leftarrow \{\theta\} - \alpha \nabla_{\theta} J$
\ENDFOR

\RETURN Optimized vine state $\{\theta\}$
\end{algorithmic}
\end{algorithm}

\subsection{Actuator Design Synthesis}
\label{design_synthesis}

Although it is desirable to parameterize actuators along the vine in terms of their \newtext{produced} turning angle $\theta_\text{curv}$, actuators are actually parameterized by pressure and length $P_\text{act}$ and $l_0$, for simulation and fabrication.
Finding these two parameters from $\theta_\text{curv}$ is nontrivial, since multiple solutions of $P_\text{act}$ and $l_0$ can satisfy any given curvature $\theta_\text{curv}$.  
We address this
\newtext{redundancy} by isolating the specific portion of the sPAM model that leads to this non-uniqueness and introducing a cost term to favor an unambiguous single solution out of the space of all possible $P_\text{act}$ and $l_0$. 

First, we solve for the actuator length $l_0$ and strain $\varepsilon$ that satisfy the target curvature $\theta_\text{curv}$ as specified by~\cref{wrinkling_model} and~\cref{eq:sPAM_force}. 
Second, we compute the corresponding parameters $\phi_{Rc}$ and $m$ by solving the nonlinear system in~\cref{eq:sPAM_model} using a gradient-based root-finding method; the Levenberg-Marquardt (LM) method~\cite{lm} proves effective, achieving convergence within $1\times 10^{-6}$ tolerance.
To avoid local optima inherent in gradient methods, we initialize the LM solver with the lowest-cost candidate from 1000 randomly sampled starting points.
Each root-finding iteration requires evaluation of Legendre elliptic integrals of the first and second kind, which lack analytical forms and are approximated to $1\times 10^{-10}$ precision using 10 expansions of the Taylor series.
This two-stage approach eliminates the one-to-many relationships exhibited in the first step, ensuring that a single solution is found in both stages of the actuator synthesis process.

To resolve 
\newtext{redundancies}, we impose a cost function that favors higher pressure $P_\text{act}$ over larger actuator length $l_0$, except where such preferences would yield infeasible solutions. 
When multiple valid combinations of $P_\text{act}$ and $l_0$ exist, we select the design with the highest pressure from the feasible set. 
This selection criterion is motivated by experimental observations that higher pressures produce more consistent curvatures despite fabrication tolerances and modeling uncertainties. 
Although the parameter sets $\{l_0\}$ and $\{P_\text{act}\}$ form continuous one-dimensional manifolds in the design space, we restrict sampling to discrete values due to physical fabrication constraints.
Specifically, our fabrication process is limited by a 1mm accuracy for $l_0$ and we are only capable of providing pressures from a discrete set, which limits the available parameter values. 
This process ultimately yields optimal pressure $P_\text{act}^*$ and corresponding actuator length $l_0^*$.
We also note that throughout this work, we maintain constant values for several parameters that were excluded from the notation: constriction radius $R_c = 5$mm, actuator tube radius $R_\text{act} = 17.18$mm, vine segment radius $R_\text{vine} = 33.35$mm, and vine robot growth pressure $P_\text{vine} = 1.5$psi.

\subsection{Neural Surrogate}
\label{surrogate}

As detailed in~\cref{design_synthesis}, actuator design parameter generation requires precise solutions to the sPAM model in~\cref{eq:sPAM_model}.
While numerical methods provide reasonable performance for individual solutions (requiring only milliseconds), we wish to use our actuator model in a design optimization task, which demands hundreds of thousands of simulation steps, corresponding to millions of individual sPAM \newtext{model} evaluations.
In this high-throughput regime, numerical methods become computationally prohibitive.
We instead approximate the sPAM model using a neural network surrogate.
Analysis of the function reveals smooth and monotonic behavior of the output with respect to input parameters.
The absence of high-frequency features and the single-mode nature of the function make accurate approximation by a neural surrogate possible, as such networks excel at fitting smooth, well-behaved data.
Furthermore, neural networks can be evaluated efficiently in parallel.

\begin{figure}[H]
    \centering
    \includegraphics[width=0.50\linewidth]{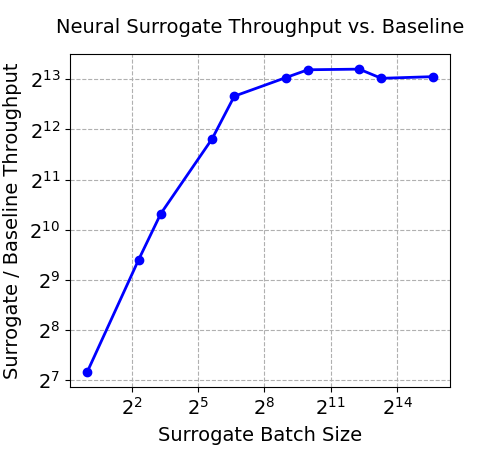}\hspace{-10pt}
    \includegraphics[width=0.50\linewidth]{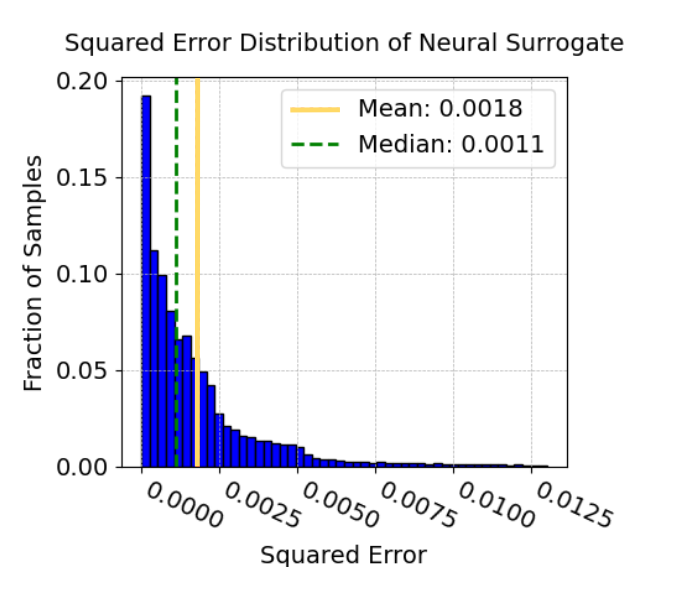}
    \caption{\textbf{Left:} The neural surrogate demonstrates a major speed up over the baseline unbatched method, especially as the batch size increases. Batched GPU processing can greatly accelerate parallelizable simulation tasks.
    \textbf{Right:} Distribution of neural surrogate squared error as compared with ground truth numeric method. 
    We uniformly sampled 40,000 values over the surrogate's input space and normalized the outputs.
    }
    \label{fig:nntimes}
\end{figure}

\newtext{The input parameters for the neural surrogate are sPAM actuator strain $\epsilon$ and actuator length $l_0$, and the outputs are geometric parameters $m$ and $\phi_{Rc}$. This is illustrated as step 3 in \cref{alg:sim_step}. sPAM parameters $m$ and $\phi_{Rc}$ are required to calculate actuator force $F_{t}$ in \cref{eq:sPAM_force}. This is step 4 in \cref{alg:sim_step}. The training data for neural surrogate is generated by uniformly sampling sPAM actuator parameters $m \in (0, 0.5]$ and $l_0\in [0.02, 0.08]$m and running them through our full analytical model from \cref{eq:sPAM_model} to get the corresponding values of $\phi_{Rc} \in (0, \frac{\pi}{2}]$ and $\epsilon \in [0, 1]$.} Through hyperparameter optimization, we determined that a $2\times32\times2$ multi-layer perceptron with ReLU nonlinearities~\cite{nair2010rectified} and He initialization~\cite{he2015delving} provided the best balance of speed and accuracy. We trained the surrogate \newtext{with a 80/10/10 train/test/validation split} on 40,000 ground truth samples. 
Performance evaluation results are presented in~\cref{fig:nntimes}.
The surrogate achieves four orders of magnitude improvement in throughput (at experimental batch sizes) while maintaining average MSE loss of 0.003 (with outputs normalized).
These computational speeds enable simulation faster than real-time, making our design optimization tractable for complex environments.

\section{Long Horizon Design Optimization}\label{design_optimization}

In contact-rich environment navigation problems, contacts, nonlinear bending mechanics, and growth dynamics jointly determine vine behavior.
Analytical or inverse numerical solutions for vine design under these coupled effects are either intractable or computationally challenging, necessitating approaches that use forward simulation.
We formulate the design optimization problem for soft growing robots as a sampling-based kinodynamic planning (SBKP) problem~\cite{lavalle2006planning}.
SBKPs provide a framework for solving problems that can only be explored with forward dynamic simulation---in our problem, we simulate the vine robot under varying actuator design parameters and propagate the vine robot forward in time by simulating constant growth with sampled actuator designs.

Specifically, we use \newtext{a modification to the} Stable Sparse RRT* (SST*)~\cite{li2016} \newtext{algorithm}, which provides asymptotic near-optimality guarantees \newtext{for finding dynamic paths between points given a cost}.
That is, as the number of samples drawn goes to infinity, the cost of the solution converges to $1 + \epsilon$ times the optimal solution.
\newtext{Specifically, we use} this approach \newtext{to} enable the discovery of robot designs that \newtext{minimize actuator requirements to achieve navigation goals, often by exploiting} environmental contact.

\begin{figure}[H]
    \centering
    \includegraphics[width=0.96\linewidth, trim=0 70pt 0 5pt, clip]{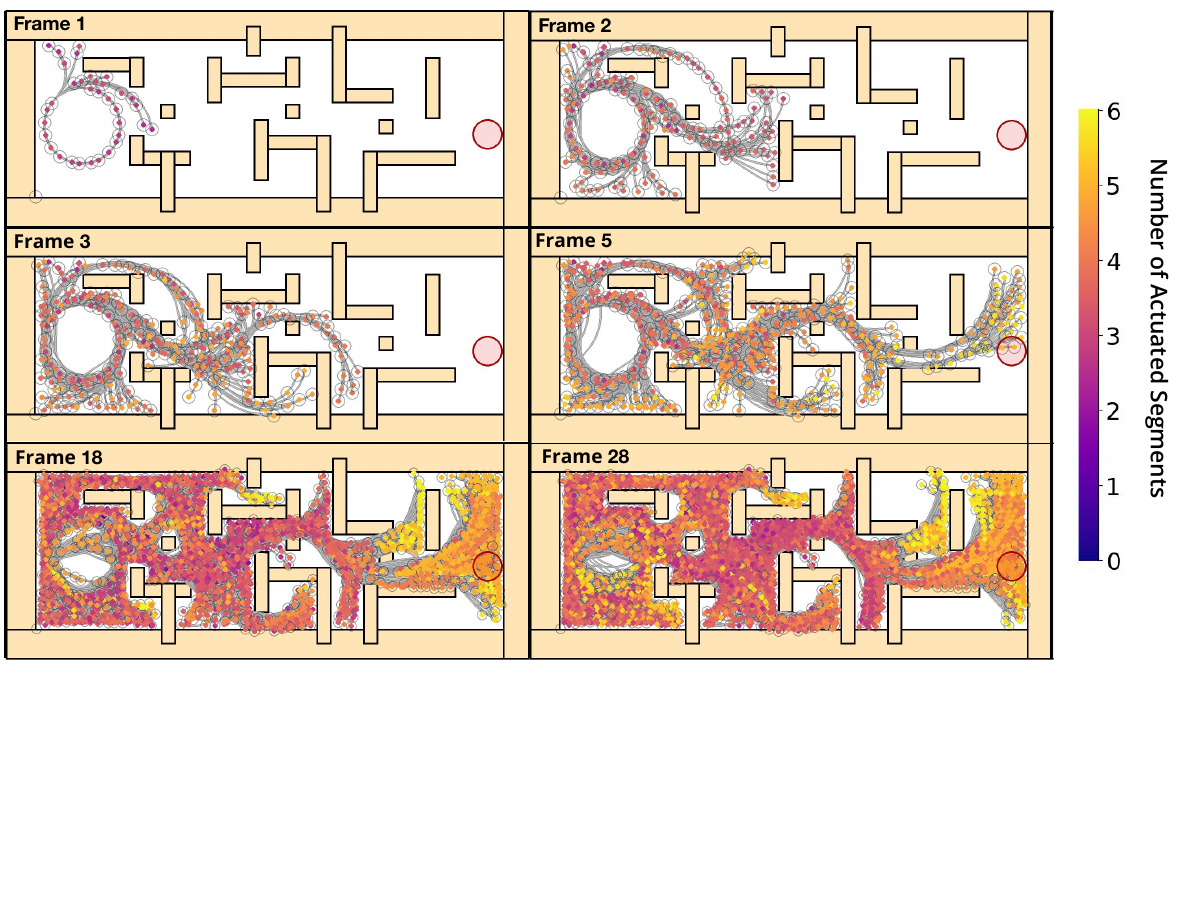}
    \caption{\newtext{Progression of planning in the \textbf{Long} environment, illustrating rapid initial exploration followed by convergent optimization.
      The planner identifies an initial solution by iteration 5, then refines the design through exploration of alternative contact-exploitation strategies. The colorbar on the right represents the number of actuated segments.}}
    \label{fig:rollout}
\end{figure}

We modify SST* to make use of the simulation framework from~\cref{sec:sim} to provide batch simulation capabilities for the planner, \newtext{expanding many nodes simultaneously, rather than one at a time.}
At each iteration, the planner chooses a varying time of execution and actuation parameters to propagate, sampling from existing nodes in the search tree to grow from---the nodes in the search tree are parameterized by the pose of the tip of the vine in space.
An example of how the planner progresses its search with these batch rollouts is shown in~\cref{fig:rollout}.

\subsection{Reverse Tree Heuristic}
To reduce SST*'s computational cost, we incorporate a reverse tree search heuristic that uses geometric (not dynamic) rollouts of designs to guide exploration toward goal-reaching areas. 
Cost-to-go heuristics improve search efficiency in methods like AIT*~\cite{aitstar} and GBRRT~\cite{ottegbrrt}, which employ bidirectional trees where the reverse tree provides guidance rather than direct connection.
Our reverse tree uses RRT*~\cite{rrtstar} in OMPL~\cite{ompl} with biarc interpolation, providing asymptotic optimality. 
Biarcs are two tangentially continuous circular arcs and align well with the piecewise constant curvature behavior of actuated vine segments~\cite{wang22}. 
We modify SST*'s cost function to include both cost-to-come and estimated cost-to-go from the reverse geometric tree, which establishes upper bounds on path costs used by the kinodynamic planner.
The objective function minimizes the number of curved segments (actuator count) with path length as a tie-breaker.
For state validity checking, we verify that the curvature of sampled biarcs $\theta_\text{curv}$ lies within numerically determined feasible minimum and maximum values rather than computing explicit actuator designs, maintaining planning soundness while reducing computation.

\newtext{\subsection{Planner Algorithm} 
The details of the planner are shown in \cref{alg:sststar}. The planner is initialized with an empty tree data structure $\mathcal{T}$ and a predetermined start state $x_{\text{start}}$, goal region $\mathcal{X}_{\text{goal}}$, obstacle list $\mathcal{X}_{\text{obs}}$, and batch size $B$. 
States in the planner include a full robot state $\{\theta\}$ and the pose of the tip of the robot $p \in SE(2)$.
The start state is inserted into $\mathcal{T}$. 
Next, $B$ states are selected for expansion $x_{\text{exp}}$, prioritizing states with the lowest cost. This cost is the sum of the cost of actions taken to reach the state and the cost-to-goal heuristic of the nearest node in the reverse geometric tree. In the first iteration, only the start state can be expanded. The planner then samples random bending controls and the number of timesteps $T$ to propagate. After repeating the simulation step, \cref{alg:sim_step}, $T$ times, resultant states $x_{\text{res}}$ are found. However, because there are $B$ states, with many overlapping, a pruning step is necessary to maintain optimality.
States are ``overlapping'' if the robot's tip is within a diminishing radius $r$ according to an $SE(2)$ metric.
This radius is reduced by a decay factor after certain iterations determined by a scheduling function; we choose a value of $60$ for the initial radius (in the $SE(2)$ metric space, which combines millimeters and angular distance) and reduce by multiplying by $\alpha = 0.8$.
Among the overlapping states, the planner keeps the state with the lowest cost and prunes the others to maintain the sparsity of the tree. The remaining states after pruning are inserted into $\mathcal{T}$. Finally, the planner checks if any of the inserted states fall within the goal region. If the goal is not reached, the planner reselects states to expand for the next iteration.

Once the goal has been reached, the solution path can be found by following the parents of the leaf up to the root of the tree, which is the start state. Because the edges of the tree represent actions taken to go from the parent to child state, we can always find the series of actions needed to traverse from the starting state to any other state in the tree.
}
\begin{algorithm}
\caption{\newtext{Batched SST*}}
\label{alg:sststar}
\begin{algorithmic}[1]

\STATE \textbf{Input:} $x_{\text{start}}, \mathcal{X}_{\text{goal}}, \mathcal{X}_{\text{obs}}, B, r_\text{init}$
\STATE \textbf{Initialize:} tree $\mathcal{T} \leftarrow \{x_{\text{start}}\}$, solution $\mathcal{X}_{\text{sol}} \leftarrow \{\}$, radius $r \leftarrow r_\text{init}$

\STATE $i \leftarrow 1$, $j \leftarrow 1$, $\text{sst\_iter}_0 \leftarrow 7$
\STATE $\text{next\_update\_iter} \leftarrow \text{sst\_iter}_0$

\WHILE{not isGoal($\mathcal{T}, \mathcal{X}_{\text{goal}}$)}
    \STATE $x_{\text{exp}}^0 \leftarrow \text{sample}(B, \mathcal{T})$ \label{alg:batch_sample}
    \STATE $\epsilon, l_0, T \leftarrow \text{sample-action}(B)$ \label{alg:act_sample}
    \FOR{$t = 1$ to $T$} \label{alg:propagate}
        \STATE $x_{\text{exp}}^{t} \leftarrow \textup{\cref{alg:sim_step}}(x_{\text{exp}}^{t- 1}, \epsilon, l_0, \mathcal{X}_{\text{obs}})$
    \ENDFOR
    \STATE $x_{\text{res}} \leftarrow x_{\text{exp}}^{T}$
    \STATE $x_{\text{valid}} \leftarrow \text{prune}(x_{\text{res}}, r)$ \label{alg:prune}
    \STATE $\mathcal{T} \leftarrow \mathcal{T} \cup x_{\text{valid}}$
    \IF{$i = \text{next\_update\_iter}$}
        \STATE $r \leftarrow \alpha r$, $j \leftarrow j + 1$
        \STATE $\text{next\_update\_iter} \leftarrow i +  \text{sst\_iter}_0 \cdot (1 + \log_2(j)) \cdot \alpha^{-(d+l+1)j} $
    \ENDIF
    \STATE $i \leftarrow i + 1$
\ENDWHILE

\STATE $\mathcal{X}_{\text{sol}} \leftarrow \text{solution}(\mathcal{T})$
\RETURN $\mathcal{X}_{\text{sol}}$

\end{algorithmic}
\end{algorithm}

\section{Experimental Results}\label{sec:exp_res}

\newtext{With our simulation framework established, we now seek to} evaluate \newtext{the model and simulator} against physical vine robot behavior and assess \newtext{the effectiveness of the simulator in enabling} the design optimization framework across a number of challenging environments. \newtext{In addition, inspired by the previous analysis of vine robot planning using contact to navigate environments~\cite{greer20}, we use the simulator to evaluate how contact with the environment affects robustness of planned robot designs and how it reduces down the variation in vine robot states after obstacle contacts.}

\newtext{\subsection{Physical Validation of Simulator and Planner}\label{sec:phys_validation}}

\begin{figure}[H]
    \centering
    \includegraphics[width=\linewidth]{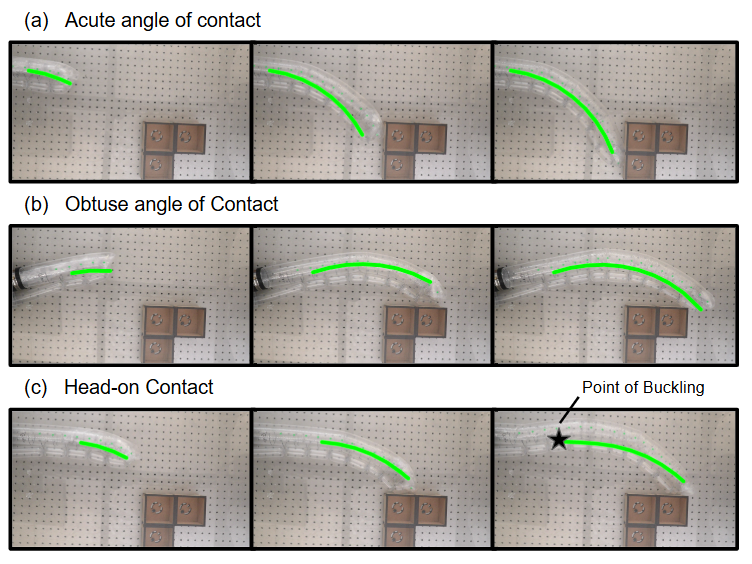}
    \caption{Single-obstacle environment experiment comparing simulated predictions for different obstacle contact conditions. 
    \textbf{(a)} Acute angle of contact, \textbf{(b)} Obtuse angle of contact, \textbf{(c)} Head-on contact. 
    The robot buckles at a point near its base. }
    \label{fig:single_obstacle}
\end{figure}

For initial validation, we constructed a single-obstacle environment to evaluate the accuracy of our contact and actuation models under well-defined conditions.
We prepared three distinct contact interaction configurations: (1) an acute angle of contact, designed to bend into the contact normal (increasing curvature under contact), (2) an obtuse angle of contact, designed to bend away from the contact normal (decreasing curvature under contact), and (3) a head-on contact, which leads to buckling behavior as the beam bends away from the contact normal (\cref{fig:single_obstacle}). \newtext{We calculate the root mean square error (RMSE) at each video frame compared to the corresponding simulator step, using the vine robot radius ($R_\text{vine}$=33.35mm) as a reference dimension when evaluating the deviations. The RMSE distributions for an acute angle of contact, an obtuse angle of contact, and a head-on contact are 20.6$\pm$3.8mm, 22.2$\pm$5.1mm, and 15.8$\pm$5.2mm, respectively.}

\begin{figure}[H]
    \centering
    \includegraphics[width=1\linewidth]{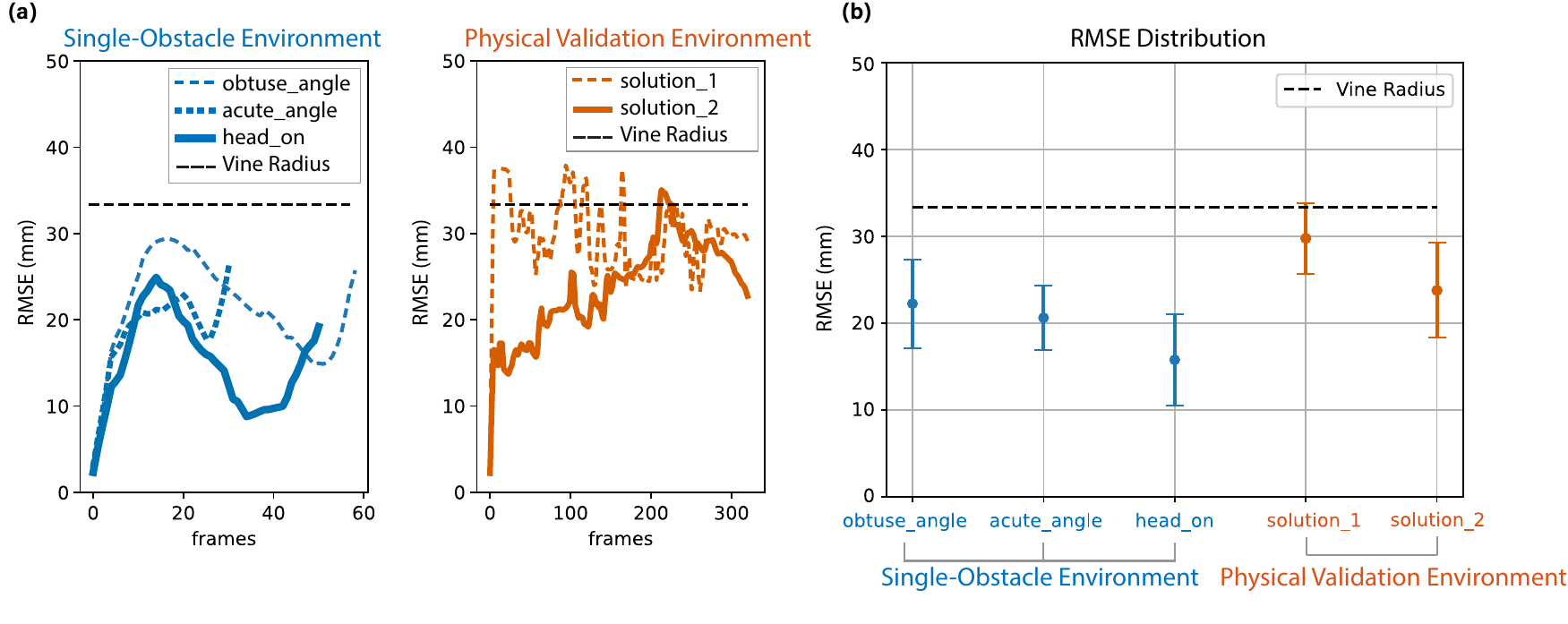}
    \caption{\newtext{(a) Frame-by-frame RMSE, (b) Distribution of RMSE between simulated predictions and deployed robot behavior for single-obstacle and physical validation environments.}}
    \label{fig:rmse_time}
\end{figure}

\newtext{Additionally, to} evaluate \newtext{the physical match of the designs produced by} our design optimization approach, we constructed a multi-obstacle environment with many potential solutions that required exploitation of contacts for successful navigation.
First, we applied our design optimization framework to find a vine robot actuator design that would navigate the robot from a specified start pose to a target goal region. 
Second, we fabricated the generated designs. 
Third, we deployed these physical robots and compared their behavior against the simulator's predictions. \newtext{The design optimization framework was able to find multiple feasible paths for our real-life demonstrations. The simulated predictions superimposed on the deployed robot in~\cref{fig:sol_1} demonstrate that our simulator accurately models both contact mechanics and bending behavior and that our generated designs translate effectively to physical implementations. Quantitative analysis of the deviation between simulated predictions and deployed robots shows RMSE of 29.8$\pm$4.1 mm for solution 1 and 23.8$\pm$5.5 mm for solution~2. 

\cref{fig:rmse_time} shows the frame-by-frame deviation between simulated predictions and deployed robot behavior for single-obstacle and multi-obstacle environments and ~\cref{tbl:RMSE_results} summarizes the results.}

\subsection{Long-Horizon Design Optimization}
\label{envs}

\begin{figure}[H]
    \centering
    \includegraphics[width=\linewidth]{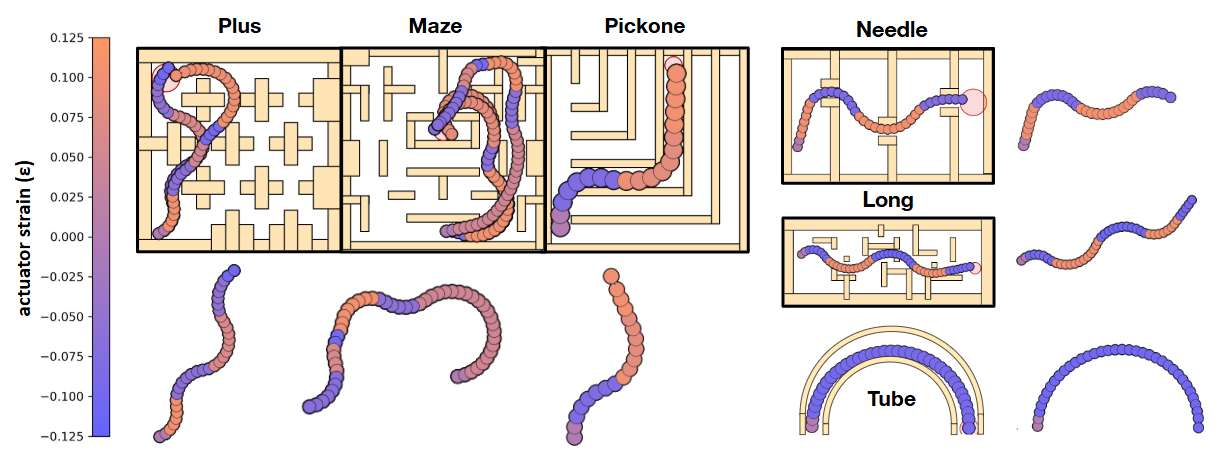}
    \caption{Multiple solutions found for the six evaluation environments by our design optimization approach. 
    Our approach successfully identified solutions for: complex navigation through clutter (\textbf{Plus}, \textbf{Maze}, and \textbf{Pickone}), navigation through tight constraints (\textbf{Needle}), precise contact exploitation (\textbf{Long}), and trivial solutions when they exist (\textbf{Tube}). 
    Free-floating images display the final actuated vine shape in free space.}
    \label{fig:envs_sol}
\end{figure}

We assess the effectiveness of our design optimization framework across six different environments, each designed to test specific aspects of the planning approach and environmental interaction capabilities.
The evaluation environments, illustrated in~\cref{fig:envs_sol}, span a range of planning challenges:
\begin{enumerate}
\item \textbf{Plus}: A uniform grid of plus-shaped obstacles creating multiple viable paths, testing the planner's ability to identify better solutions among alternatives.
\item \textbf{Maze}: A structured environment with few solutions and many dead-ends, evaluating planning robustness in constrained solution spaces.
\item \textbf{Pickone}: Multiple identical passages with only one leading to the goal, evaluating the effectiveness of heuristic guidance.
\item \textbf{Needle}: A series of narrow passages requiring precision to navigate, evaluating the planner's ability to handle tight geometric constraints.
\item \textbf{Long}: An environment with strategically positioned contacts that must be exploited to avoid dead-end passages, testing contact-based navigation.
\item \textbf{Tube}: A minimal environment to test whether the planner can identify trivial solutions.
\end{enumerate}

Our design optimization framework successfully identified feasible solutions for all evaluation environments, as demonstrated in~\cref{fig:envs_sol}.
Even in the more difficult environments such as \textbf{Maze}, \textbf{Long}, and \textbf{Needle}, which are cluttered, full of dead ends, and require precise movement, our planner effectively discovered minimum-actuator designs that exploited contact with the environment to achieve the task objective.

We evaluated the planner with and without the reverse tree heuristic over 30 independent runs on each environment; statistics are reported in~\cref{tbl:results}.
The planner had consistent performance; all environments achieved initial solutions within 60 seconds on an AMD Ryzen 9 5900 CPU and NVIDIA RTX 3050 GPU, enabling interactive design iteration and deployment with a desktop computer.

\subsection{Robustness of Designs}
\label{sec:robust}

\begin{figure}[H]
    \centering
    \includegraphics[width=0.8\linewidth]{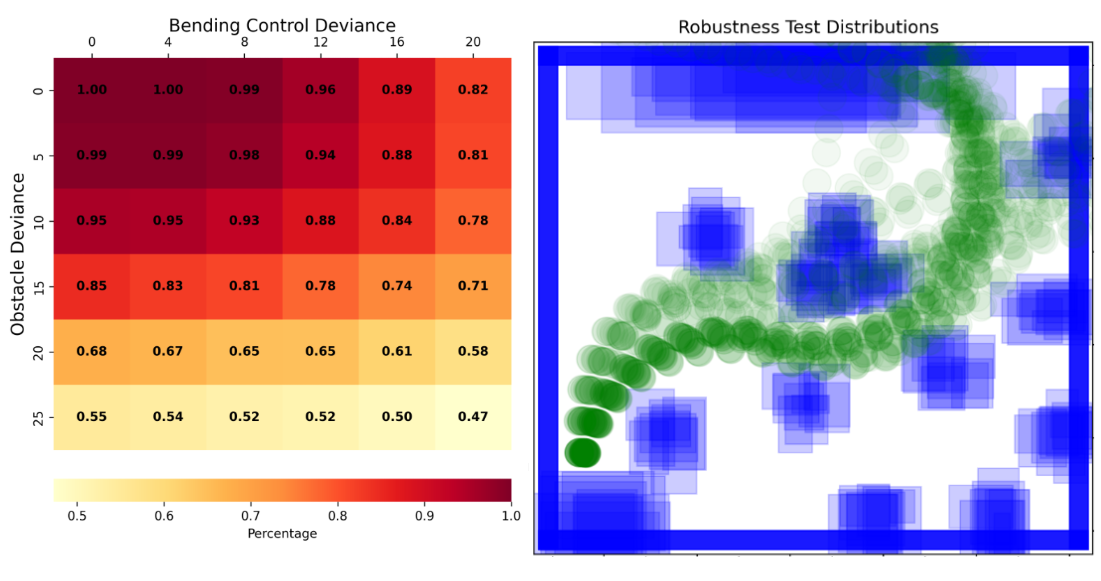}
    \caption{\newtext{\textbf{Left}: Robustness test of a nominal design over Gaussian noise applied to obstacles and produced design. Scale for obstacle shift is the percent size difference and position change. The scale for bending control shift is the percent change to actuator pressure and actuator length. Each division was run with a sample size of 1000, and the fraction of successes is reported. \textbf{Right}: Examples of 10 sampled environments at the highest obstacle deviation and 60 rollouts with varying bending control deviation from the same nominal design. The "funneling" effects of contacts are clear: even with variations in obstacles and designs, the vine robot is pushed down similar trajectories.}}
    \label{fig:robust}
\end{figure}

\cref{fig:robust} presents the success rate of reaching the goal region as a function of uncertainty in both the environment and actuation parameters, in the same environment as~\cref{fig:sol_1}.
We model environmental uncertainty through Gaussian perturbations of obstacle positions and sizes with gradually increasing variance levels.
Similarly, actuation uncertainty is modeled through perturbations of actuator pressure and length parameters.
For each level of uncertainty, we evaluate success rates across 1000 trials.

Results show graceful degradation of task satisfaction with increasing uncertainty levels, and that designs generated by our framework have inherent tolerance to uncertainties, particularly when contact interactions provide environmental guidance that compensates for actuation variations.
This supports the deployment of found designs in real-world environments where perfect transfer of simulation to reality is unattainable.

\newtext{\subsection{Funneling Behavior}\label{funnelling}
A separate study was run to investigate the vine robots' inherent tendency to display funnel-like behaviors after contact with obstacles, which causes trajectories with minor deviations to converge~\cite{greer20}. ~\cref{fig:funnel} illustrates 50 vine rollouts with uniformly distributed starting angles in 4 environments as obstacles gradually become denser. Some variance was introduced to the bending controls to bias growth towards the top right corner. Dispersion statistics were run on the endpoints of the trajectories and are displayed in ~\cref{fig:funnel_stats}.

Dispersion analysis was done using k-nearest neighbor density concentration ratio of the top 10 percent of clusters. Given an endpoint of a vine trajectory \(i\), the average distance \(r_k(i)\) between point \(i\) and its \(k\) nearest neighbors is calculated. The local density estimate around point \(i\) would be:

\[p_i = \frac{k}{\pi r_k(i)^2}\]

The larger \(p_i\) is, the denser the cluster around point \(i\). The density concentration ratio is the ratio between the average of the 90th percentile of \(p_i\) and its mean. A higher value indicates a higher degree of funneling of vine trajectories, as seen with comparing the free space results to all three environments with obstacles.

}

\section{Discussion}

 \newtext{
 This work introduces a unified modeling framework for soft growing robots, effectively capturing the interactions between pneumatic actuation, beam mechanics, and environmental contact. The innovative neural surrogate approach allows for efficient simulation of vine robots, achieving a computational speedup of four orders of magnitude compared to traditional methods. This advancement enables faster-than-real-time simulation, which is crucial for developing a contact-aware design optimization framework. In this section, we discuss the performance and limitations of our framework and highlight major advances.}

\begin{figure}[H]
    \centering
    \includegraphics[width=0.8\linewidth]{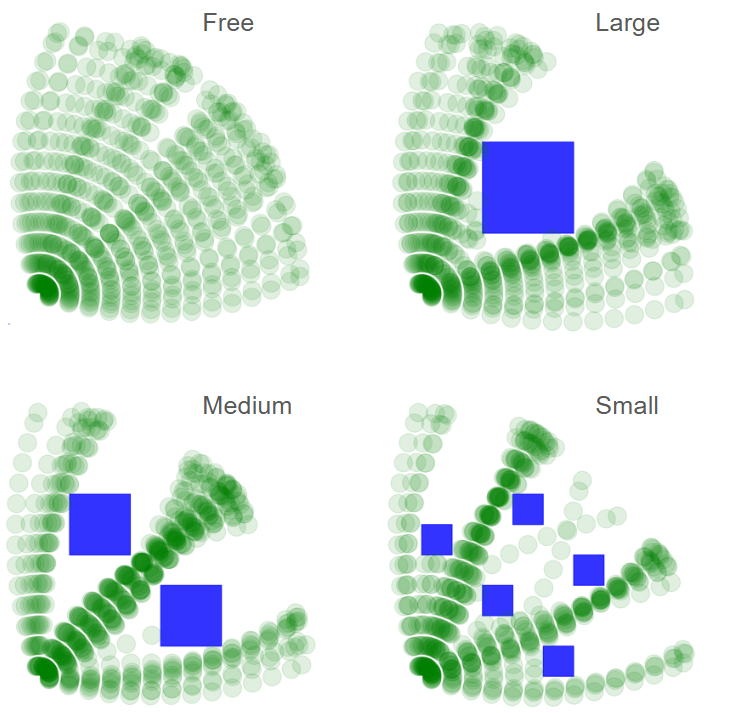}
    \caption{\newtext{Vine rollouts are run with slight deviations to start angle. The environment with no obstacles shows the even spread of vine trajectories. The other environments show the funneling behavior, forming highways where many trajectories converge.}}
    \label{fig:funnel}
\end{figure}

\begin{figure}[H]
    \centering
    \includegraphics[width=0.8\linewidth]{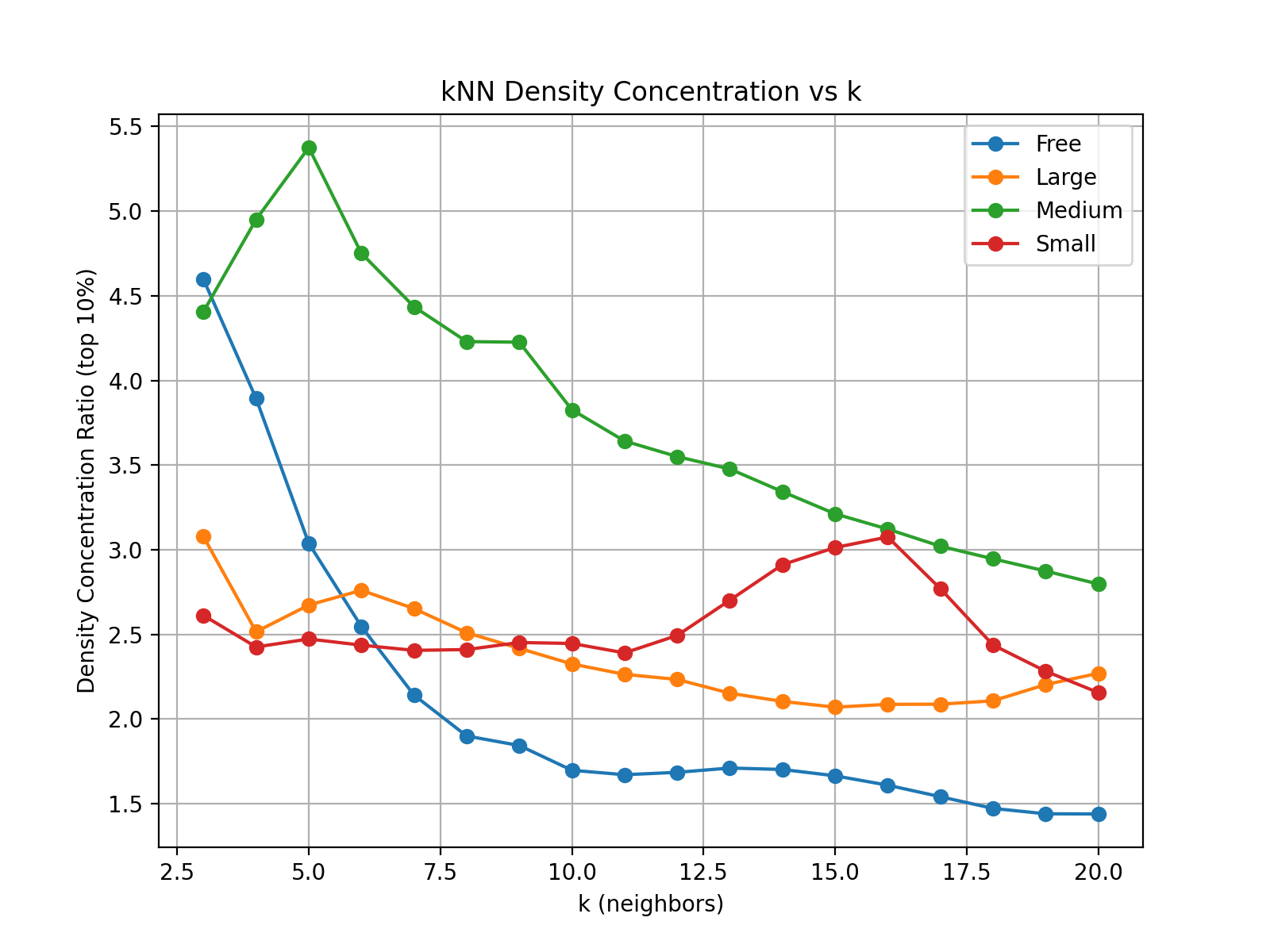}
    \caption{\newtext{K-nearest neighbors concentration squadensity ratio across multiple values of k and the 4 environments shown in Fig~\ref{fig:funnel}.}}
    \label{fig:funnel_stats}
\end{figure}

\newtext{\subsection{Capturing Actuator-Contact Interactions} 
The single-obstacle environment experiments in \cref{sec:phys_validation} demonstrate that our simulator captures the fundamental contact-actuation coupling accurately across the range of contact conditions. The simulator is able to capture both the constant curvature bends produced by free space actuation or contact in a direction of bending and the discrete buckling when actuation and contact direction are opposed.  Due to the nonlinearity of the sPAM bending model and contact mechanics, these configurations exhibit asymmetric behaviors and yield substantially different final robot configurations. This match is maintained for multi-object interactions, with only a slight increase in RMSE, likely due to the longer length between contact events.

It should be noted that small deviations in the simulator before obstacle contact, such as start position and heading angle, can cause divergence on contact, as contact direction and force often dominate and can ``funnel'' the robot into different homotopy classes of motion through the environment (e.g., hitting on one side of a obstacle versus another). This sensitivity is underscored by our funneling behavior evaluation in~\cref{funnelling}. While many neighboring trajectories seen in \cref{fig:funnel} clumped up in final tip position, starting angles hitting on either side of a obstacle corner ended up further apart after contact.}

\begin{figure}[H]
    \centering
    \includegraphics[width=1\linewidth]{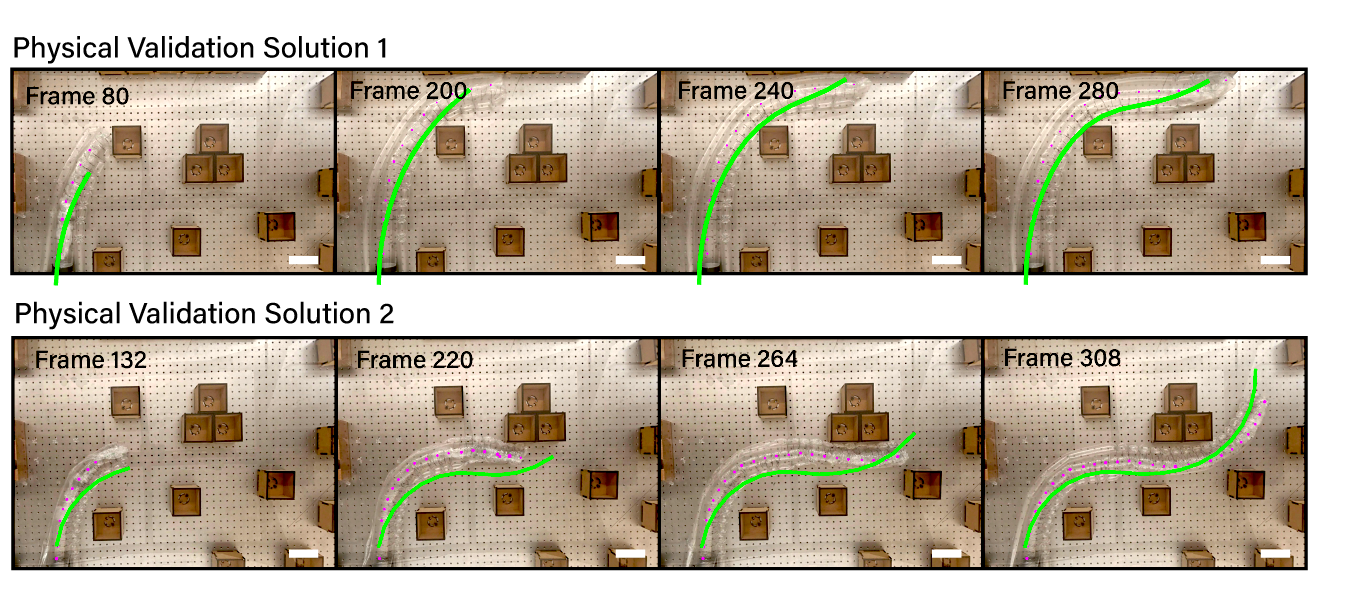}
    \caption{\newtext{Physical validation experiment comparing simulated predictions (green) with real vine robot behavior in a complex multi-obstacle environment. The designed actuator configuration successfully navigates the cluttered environment in both simulation and reality, with the mean RMSE of 29.8$\pm$4.1mm for solution 1 and 23.8$\pm$5.5mm for solution 2. The scale bar in the figure is equal to 100mm.}}
    \label{fig:sol_1}
\end{figure}

\newtext{\subsection{Application to Long-horizon Design Optimization Task}
To assess the performance of the design optimization framework, we selected six different simulated environments. These environments were explicitly designed to capture important aspects of the planner's ability, such as (1) picking a better solution from multiple alternatives, (2) handling tight geometric constraints, and (3) identifying trivial solutions. 

The \textbf{Long} environment validates the core hypothesis of our approach: the planner successfully identified designs that exploit deliberate contact interactions to navigate passages that would be inaccessible without environmental support.
This demonstrates the practical advantage of contact-aware design optimization over traditional collision-avoidance approaches.
This is also evidenced in the \textbf{Tube} environment, where the planner consistently produced near-optimal solutions with only one or two segments, indicating successful identification of minimal-actuation designs despite the availability of more complex alternatives. For environments with multiple viable solution paths (e.g., \textbf{Plus} and \textbf{Maze}), the planner demonstrated breadth of exploration, identifying diverse design alternatives that utilize different environmental corridors.
This capability provides designers with multiple implementation options with different potential robustness against environmental variations.

Surprisingly, the reverse tree heuristic did not provide substantial performance improvements across the evaluated environments, even for more complex scenarios like \textbf{Pickone} and \textbf{Maze} that feature dead-ends and branching passages where heuristic guidance would be expected to provide benefit.
We hypothesize that the natural funnel-like behavior exhibited by vine robot dynamics---particularly the tendency for contact interactions to guide the robot toward feasible solutions---reduces the need for explicit guidance in environments of the scale we evaluated.
This observation suggests that the inherent physics of vine robots implicitly guides the system, obviating the need for other heuristics.

The real-world multi-obstacle environment in \cref{fig:sol_1} shows a concrete example of the practical applications of our fast simulation framework in a contact-aware design optimization task. The planner finds designs of vine robots with minimal numbers of actuators within the constraints of feasible actuator fabrication and control. This example specifically highlights the  ability to navigate through inaccessible passages (due to limits in actuator strain) by leveraging environmental contact. The validation of the model and simulator through real robot experiments highlights their robustness, with results indicating that the mean RMSE is of the same order as the vine robot radius. Deviations of this magnitude are within acceptable tolerance for the navigation tasks considered, particularly given the inherent uncertainties in fabrication, deployment, and measurement.
}

\newtext{ \subsection{Effects of Contact on Robustness through Funneling}
The robustness tests (\cref{sec:robust}) quantified the effect of uncertainties in the environment, such as obstacle positioning, and vine robot actuation parameters. The results from the robustness tests indicate that the vine robot's behavior in an obstacle-rich environment is highly robust to these uncertainties. Only with greater than 10\% variation in obstacle or 12\% variation in bending control does the success rate drop below 90\%. Eventually, greater deviance in either environment or design will cause more significant failures, as the deviance in rollout becomes more likely to change the initial wall contacted, leading to a different homotopy of final path. The funneling behavior tests (\cref{funnelling}) showed that the vine robots indeed display funnel-like behavior that creates concentrated trajectories after contact with obstacles. These results together show that minor deviations in vine bending control have little effect on overall vine trajectory due to effects from contact forces, though this behavior shifts dramatically after a certain point.

These evaluations also reinforce our earlier observation that contact-guided navigation provides natural error correction that maintains overall success despite small deviations in behavior.
An interesting direction for future work would be to explicitly target designs that exploit behaviors that are inherently more robust to uncertainty due to this effect.}

\newtext{\subsection{Generalizing to Different Actuator and Control Models}
While we used sPAMs as active steering actuators, other actuation methods~\cite{du2023finite, kubler2024comparison}, such as the serial pouch motor, the cylindrical pneumatic artificial muscle (cPAM), the fabric pneumatic artificial muscle (fPAM), and tip steering, can be used in our simulator due to the modular feature of our framework. The neural surrogate approach used for sPAM model integration can be easily replaced with similar neural surrogate or first principle models for cPAMs and pouch motors. 

The current framework is also implemented as an open-loop control task, where the vine robot design (vine robot length, actuator placement, and actuation pressures) is selected during fabrication and remains fixed throughout the deployment. However, this can be easily extended to enable time-dependent or position-dependent steering change. Through initial simulation examples in the supplementary material, we showed how tip steering could be added to the modeling framework by temporarily penalizing growth and changing steering command.}

 \newtext{ \subsection{Limitations}
Despite these promising results, the current simulator and planner have limitations. It is confined to planar environments and relies on several assumptions about pressure dynamics which become less realistic with longer robot growth. The current framework does not take into account the mass and inertia of the vine robot, thus limiting its use for vine robots without a tip mass or tip-based retraction mechanism~\cite{coad2020retraction}. For the planner, due to its planning-as-design approach, it is currently unable to generate solution that generalize to multiple environments. While this is not an inherent limit, more work would likely be needed to expand the actuator design space to perform multi-objective design optimization. Additionally, the design framework was tested in relatively simple scenarios due to constraints on the total length of the robot in the physical experiment in order to maintain a consistent tail tension and therefore growth pressure, limiting its ability to demonstrate long-horizon guidance through geometric heuristics. Longer lengths of vine robots tend to have internal friction not currently addressed by the constant growth pressure assumption~\cite{blumenschein2017modeling}. In practice, pressurized base stations offer better control over vine robot pressure over long lengths, so future work will look to integrate the planner designs with the closed loop control of growth rate enabled by a pressurized base station. A future model would need to consider the effect of tail tension to simulate controlled growth and retraction. }

\newtext{

\section{Conclusion}

The unified modeling framework and fast simulation approach introduced in this work significantly advance the field of soft growing robots, allowing for fast, efficient, and accurate simulations. The contact-aware design optimization framework provides a novel method for creating robust vine robot designs to navigate a known environment. More research is needed to address the current limitations, such as the model’s restriction to planar environments and the simplicity of the tested scenarios. Additionally, exploring cost functions that utilize the characterized funneling effects of contact forces could enhance robustness and reduce uncertainty. Another interesting area for research could be the integration of sensing modules in the simulation framework, to enable real-time feedback-based closed-loop control of vine robots.}

\newtext{
Our work opens new avenues for exploration in addressing the challenges of soft robot path planning and control. A natural extension of this framework is to tackle multi-objective optimization tasks involving high-level reasoning and long-horizon navigation, such as payload deployment, manipulation of objects, and navigation in unknown environments. Ultimately, the advances in simulation frameworks will be the basis on which intelligent, autonomous vine robots are built.}

\section*{Acknowledgments}
We would like to thank Gilbert Chang for feedback on this manuscript.

\section*{Authorship Contribution Statement}

YG and LC developed and evaluated the simulation and planning framework and assisted in physical validation.
PB and SW developed the combined actuation and contact moment model and physically validated the approach.
ZK and LHB supervised the project.
All authors discussed the results and contributed to the final manuscript.

\section*{Conflict of Interest Statement}
The authors have no conflicts of interest to declare.

\section*{Funding Statement}
This work was supported in part by NSF FRR 2308653. This material is based upon work supported by the United States Air Force (AFMC AFRL/RXNW) under Air Force Contract No. FA2394-24-C-B060.

\bibliographystyle{plain}
\bibliography{references}

\begin{thebibliography}{10}

\bibitem{bender2014survey}
Jan Bender, Matthias M{\"u}ller, Miguel~A Otaduy, Matthias Teschner, and Miles Macklin.
\newblock A survey on position-based simulation methods in computer graphics.
\newblock {\em Computer Graphics Forum}, 33(6):228--251, 2014.

\bibitem{berthet2021mammobot}
Pierre Berthet-Rayne, SM~Hadi Sadati, Georgios Petrou, Neel Patel, Stamatia Giannarou, Daniel~Richard Leff, and Christos Bergeles.
\newblock Mammobot: A miniature steerable soft growing robot for early breast cancer detection.
\newblock {\em IEEE Robotics and Automation Letters}, 6(3):5056--5063, 2021.

\bibitem{blumenschein22}
Laura~H. Blumenschein, Margaret Koehler, Nathan~S. Usevitch, Elliot~Wright Hawkes, D.~Caleb Rucker, and Allison~M. Okamura.
\newblock Geometric solutions for general actuator routing on inflated-beam soft growing robots.
\newblock {\em IEEE Transactions on Robotics}, 38(3):1820--1840, 2022.

\bibitem{blumenschein2017modeling}
Laura~H Blumenschein, Allison~M Okamura, and Elliot~W Hawkes.
\newblock Modeling of bioinspired apical extension in a soft robot.
\newblock In {\em Conference on Biomimetic and Biohybrid Systems}, pages 522--531, 2017.

\bibitem{borvorntanajanya2024model}
Korn Borvorntanajanya, Shen Treratanakulchai, Ferdinando~Rodriguez y~Baena, and Enrico Franco.
\newblock Model-based tracking control of a soft growing robot for colonoscopy.
\newblock {\em IEEE Transactions on Medical Robotics and Bionics}, 2024.

\bibitem{bradbury2018jax}
James Bradbury, Roy Frostig, Peter Hawkins, Matthew~James Johnson, Chris Leary, Dougal Maclaurin, George Necula, Adam Paszke, Jake Vander{P}las, Skye Wanderman-{M}ilne, and Qiao Zhang.
\newblock {JAX}: composable transformations of {P}ython+{N}um{P}y programs, 2018.

\bibitem{robosoft_diff_sim}
Lucas Chen, Yitian Gao, Sicheng Wang, Francesco Fuentes, Laura~H. Blumenschein, and Zachary Kingston.
\newblock Physics-grounded differentiable simulation for soft growing robots.
\newblock In {\em IEEE International Conference on Soft Robotics}, pages 1--8, 2025.

\bibitem{coad2019vine}
Margaret~M Coad, Laura~H Blumenschein, Sadie Cutler, Javier A~Reyna Zepeda, Nicholas~D Naclerio, Haitham El-Hussieny, Usman Mehmood, Jee-Hwan Ryu, Elliot~W Hawkes, and Allison~M Okamura.
\newblock Vine robots.
\newblock {\em IEEE Robotics and Automation Magazine}, 27(3):120--132, 2019.

\bibitem{coad2020retraction}
Margaret~M Coad, Rachel~P Thomasson, Laura~H Blumenschein, Nathan~S Usevitch, Elliot~W Hawkes, and Allison~M Okamura.
\newblock Retraction of soft growing robots without buckling.
\newblock {\em IEEE Robotics and Automation Letters}, 5(2):2115--2122, 2020.

\bibitem{comer_levy}
R.~L. Comer and Samuel Levy.
\newblock Deflections of an inflated circular-cylindrical cantilever beam.
\newblock {\em AIAA Journal}, 1(7):1652--1655, 1963.

\bibitem{daerden99PPAM}
Frank Daerden.
\newblock {\em Conception and realization of pleated pneumatic artificial muscles and their use as compliant actuation elements}.
\newblock {Ph.D.} thesis, Vrije Universiteit Brussel, 1999.

\bibitem{donald1993}
Bruce Donald, Patrick Xavier, John Canny, and John Reif.
\newblock Kinodynamic motion planning.
\newblock {\em Journal of the ACM}, 40(5):1048–1066, November 1993.

\bibitem{du2023finite}
Cosima du~Pasquier, Sehui Jeong, and Allison~M Okamura.
\newblock Finite element modeling of pneumatic bending actuators for inflated-beam robots.
\newblock {\em IEEE Robotics and Automation Letters}, 2023.

\bibitem{el2018development}
Haitham El-Hussieny, Usman Mehmood, Zain Mehdi, Sang-Goo Jeong, Muhammad Usman, Elliot~W Hawkes, Allison~M Okarnura, and Jee-Hwan Ryu.
\newblock Development and evaluation of an intuitive flexible interface for teleoperating soft growing robots.
\newblock In {\em IEEE/RSJ International Conference on Intelligent Robots and Systems}, pages 4995--5002, 2018.

\bibitem{faure2012sofa}
Fran{\c{c}}ois Faure, Christian Duriez, Herv{\'e} Delingette, J{\'e}r{\'e}mie Allard, Benjamin Gilles, St{\'e}phanie Marchesseau, Hugo Talbot, Hadrien Courtecuisse, Guillaume Bousquet, Igor Peterlik, et~al.
\newblock {SOFA}: A multi-model framework for interactive physical simulation.
\newblock {\em Soft Tissue Biomechanical Modeling for Computer Assisted Surgery}, pages 283--321, 2012.

\bibitem{fichter66}
WB~Fichter.
\newblock {\em A theory for inflated thin-wall cylindrical beams}, volume D-3466.
\newblock National Aeronautics and Space Administration, 1966.

\bibitem{frias_miranda23}
Eugenio Frias-Miranda, Alankriti Srivastava, Sicheng Wang, and Laura~H. Blumenschein.
\newblock Vine robot localization via collision.
\newblock In {\em IEEE/RSJ International Conference on Intelligent Robots and Systems}, pages 2515--2521, 2023.

\bibitem{fuentes23}
Francesco Fuentes and Laura~H. Blumenschein.
\newblock Mapping unknown environments through passive deformation of soft, growing robots.
\newblock In {\em IEEE/RSJ International Conference on Intelligent Robots and Systems}, pages 2522--2527, 2023.

\bibitem{fuentes25}
Francesco Fuentes, Serigne Diagne, Zachary Kingston, and Laura~H. Blumenschein.
\newblock Exteroception through proprioception: Sensing through improved contact modeling for soft growing robots.
\newblock {\em International Journal of Robotics Research}, 2025.
\newblock Under Review.

\bibitem{girerd2024material}
C{\'e}dric Girerd, Anna Alvarez, Elliot~W Hawkes, and Tania~K Morimoto.
\newblock Material scrunching enables working channels in miniaturized vine-inspired robots.
\newblock {\em IEEE Transactions on Robotics}, 2024.

\bibitem{InchIGRAB}
Ayush Giri, Cédric Girerd, Jacobo Cervera-Torralba, Michael~T. Tolley, and Tania~K. Morimoto.
\newblock Inchigrab: An inchworm-inspired guided retraction and bending device for vine robots during colonoscopy.
\newblock {\em IEEE/ASME Transactions on Mechatronics}, pages 1--12, 2025.

\bibitem{greer20}
Joseph~D Greer, Laura~H Blumenschein, Ron Alterovitz, Elliot~W Hawkes, and Allison~M Okamura.
\newblock Robust navigation of a soft growing robot by exploiting contact with the environment.
\newblock {\em The International Journal of Robotics Research}, 39(14):1724--1738, 2020.

\bibitem{greer2017series}
Joseph~D Greer, Tania~K Morimoto, Allison~M Okamura, and Elliot~W Hawkes.
\newblock Series pneumatic artificial muscles (spams) and application to a soft continuum robot.
\newblock In {\em IEEE International Conference on Robotics and Automation}, pages 5503--5510, 2017.

\bibitem{greer19}
Joseph~D. Greer, Tania~K. Morimoto, Allison~M. Okamura, and Elliot~W. Hawkes.
\newblock A soft, steerable continuum robot that grows via tip extension.
\newblock {\em Soft Robotics}, 6(1):95--108, 2019.

\bibitem{haggerty2023control}
David~A Haggerty, Michael~J Banks, Ervin Kamenar, Alan~B Cao, Patrick~C Curtis, Igor Mezi{\'c}, and Elliot~W Hawkes.
\newblock Control of soft robots with inertial dynamics.
\newblock {\em Science Robotics}, 8(81):eadd6864, 2023.

\bibitem{haggerty19}
David~A. Haggerty, Nicholas~D. Naclerio, and Elliot~W. Hawkes.
\newblock Characterizing environmental interactions for soft growing robots.
\newblock In {\em IEEE/RSJ International Conference on Intelligent Robots and Systems}, pages 3335--3342, 2019.

\bibitem{Hauser2016}
Kris Hauser and Yilun Zhou.
\newblock Asymptotically optimal planning by feasible kinodynamic planning in a state–cost space.
\newblock {\em IEEE Transactions on Robotics}, 32(6):1431--1443, 2016.

\bibitem{hawkes2017soft}
Elliot~W Hawkes, Laura~H Blumenschein, Joseph~D Greer, and Allison~M Okamura.
\newblock A soft robot that navigates its environment through growth.
\newblock {\em Science Robotics}, 2(8), 2017.

\bibitem{he2015delving}
Kaiming He, Xiangyu Zhang, Shaoqing Ren, and Jian Sun.
\newblock Delving deep into rectifiers: Surpassing human-level performance on imagenet classification.
\newblock In {\em IEEE International Conference on Computer Vision}, pages 1026--1034, 2015.

\bibitem{hwee23}
Joel Hwee, Andrew Lewis, Allison Raines, and Blake Hannaford.
\newblock Kinematic modeling of a soft everting robot from inflated beam theory.
\newblock In {\em IEEE International Conference on Soft Robotics}, pages 1--6, 2023.

\bibitem{jitosho2021dynamics}
Rianna Jitosho, Nathaniel Agharese, Allison Okamura, and Zac Manchester.
\newblock A dynamics simulator for soft growing robots.
\newblock In {\em IEEE International Conference on Robotics and Automation}, pages 11775--11781, 2021.

\bibitem{endoscope}
Nam~Gyun Kim, Shinwoo Park, Dongoh Seo, Sanghun Lee, Hyuk Yoon, Jaihwan Kim, and Jee-Hwan Ryu.
\newblock A soft growing robotic endoscope for painless and strain-free insertion.
\newblock {\em Soft Robotics}, 0(0):null, 0.
\newblock PMID: 40833842.

\bibitem{kubler19}
Alexander~M. K\"{u}bler, Cosima du~Pasquier, Andrew Low, Betim Djambazi, Nicolas Aymon, Julian F\"{o}rster, Nathaniel Agharese, Roland Siegwart, and Allison~M. Okamura.
\newblock A comparison of pneumatic actuators for soft growing vine robots.
\newblock {\em Soft Robotics}, 11(5):857--868, 2024.

\bibitem{kubler2024comparison}
Alexander~M K{\"u}bler, Cosima Du~Pasquier, Andrew Low, Betim Djambazi, Nicolas Aymon, Julian F{\"o}rster, Nathaniel Agharese, Roland Siegwart, and Allison~M Okamura.
\newblock A comparison of pneumatic actuators for soft growing vine robots.
\newblock {\em Soft Robotics}, 2024.

\bibitem{lavalle2006planning}
S.M. LaValle.
\newblock {\em Planning Algorithms}.
\newblock Cambridge University Press, 2006.

\bibitem{li2021bioinspired}
Pengchun Li, Yongchang Zhang, Guangyu Zhang, Dekai Zhou, and Longqiu Li.
\newblock A bioinspired soft robot combining the growth adaptability of vine plants with a coordinated control system.
\newblock {\em Research}, 2021.

\bibitem{li2016}
Yanbo Li, Zakary Littlefield, and Kostas~E. Bekris.
\newblock Asymptotically optimal sampling-based kinodynamic planning.
\newblock {\em The International Journal of Robotics Research}, 35(5):528--564, 2016.

\bibitem{xpbd}
Miles Macklin, Matthias Müller, and Nuttapong Chentanez.
\newblock {XPBD}: Position-based simulation of compliant constrained dynamics.
\newblock In {\em International Conference on Motion in Games}, page 49–54, 10 2016.

\bibitem{lm}
Jorge~J. Mor{\'e}.
\newblock The {L}evenberg-{M}arquardt algorithm: Implementation and theory.
\newblock In G.~A. Watson, editor, {\em Numerical Analysis}, pages 105--116, Berlin, Heidelberg, 1978. Springer Berlin Heidelberg.

\bibitem{nair2010rectified}
Vinod Nair and Geoffrey~E Hinton.
\newblock Rectified linear units improve restricted boltzmann machines.
\newblock In {\em International Conference on Machine Learning}, pages 807--814, 2010.

\bibitem{ottegbrrt}
Sharan Nayak and Michael~W. Otte.
\newblock Bidirectional sampling based search without two point boundary value solution.
\newblock {\em CoRR}, abs/2010.14692, 2020.

\bibitem{10844318}
Nicolas Perrault, Qi~Heng Ho, and Morteza Lahijanian.
\newblock Kino-{PAX}: Highly parallel kinodynamic sampling-based planner.
\newblock {\em IEEE Robotics and Automation Letters}, 10(3):2430--2437, 2025.

\bibitem{qin2024design}
Yimeng Qin, Allison~M Okamura, and Marco Salathe.
\newblock Design, control, and modeling for soft growing robot deployment for nuclear material inspection.
\newblock In {\em Institute of Nuclear Materials Management Annual Meeting}, 2024.

\bibitem{selvaggio20}
M.~Selvaggio, L.~A. Ramirez, N.~D. Naclerio, B.~Siciliano, and E.~W. Hawkes.
\newblock An obstacle-interaction planning method for navigation of actuated vine robots.
\newblock In {\em IEEE International Conference on Robotics and Automation}, pages 3227--3233, 2020.

\bibitem{colonoscope}
Jialei Shi, Korn Borvorntanajanya, Kaiwen Chen, Enrico Franco, and Ferdinando Rodriguez~y Baena.
\newblock Design, control, and evaluation of a novel soft everting robot for colonoscopy.
\newblock {\em IEEE Transactions on Robotics}, 41:4843--4859, 2025.

\bibitem{rrtstar}
Kiril Solovey, Lucas Janson, Edward Schmerling, Emilio Frazzoli, and Marco Pavone.
\newblock Revisiting the asymptotic optimality of {RRT}.
\newblock In {\em IEEE International Conference on Robotics and Automation}, pages 2189--2195, 2020.

\bibitem{aitstar}
Marlin~P. Strub and Jonathan~D. Gammell.
\newblock Adaptively informed trees ({AIT*}): Fast asymptotically optimal path planning through adaptive heuristics.
\newblock In {\em IEEE International Conference on Robotics and Automation}, 2020.

\bibitem{ompl}
Ioan~A. {\c{S}}ucan, Mark Moll, and Lydia~E. Kavraki.
\newblock The {O}pen {M}otion {P}lanning {L}ibrary.
\newblock {\em IEEE Robotics and Automation Magazine}, 19(4):72--82, December 2012.
\newblock \url{https://ompl.kavrakilab.org}.

\bibitem{vartholomeos2024lumped}
Panagiotis Vartholomeos, Zicong Wu, SM~Hadi Sadati, and Christos Bergeles.
\newblock Lumped parameter dynamic model of an eversion growing robot: Analysis, simulation and experimental validation.
\newblock In {\em IEEE International Conference on Robotics and Automation}, pages 12734--12740, 2024.

\bibitem{wang22}
Sicheng Wang and Laura~H. Blumenschein.
\newblock A geometric design approach for continuum robots by piecewise approximation of freeform shapes.
\newblock In {\em IEEE/RSJ International Conference on Intelligent Robots and Systems}, pages 5416--5423, 2022.

\bibitem{sPAM}
Sicheng Wang and Laura~H. Blumenschein.
\newblock Refined modeling for serial pneumatic artificial muscles enables model-based actuation design.
\newblock In {\em IEEE International Conference on Soft Robotics}, pages 800--807, 2024.

\bibitem{wang2024anisotropic}
Sicheng Wang, Eugenio Frias-Miranda, Antonio~Alvarez Valdivia, and Laura~H Blumenschein.
\newblock Anisotropic stiffness and programmable actuation for soft robots enabled by an inflated rotational joint.
\newblock {\em arXiv Preprint arXiv:2410.13003}, 2024.

\bibitem{webb2013}
Dustin~J. Webb and Jur van~den Berg.
\newblock Kinodynamic {RRT*}: Asymptotically optimal motion planning for robots with linear dynamics.
\newblock In {\em IEEE International Conference on Robotics and Automation}, pages 5054--5061, 2013.

\bibitem{wu2023towards}
Zicong Wu, Mikel De~Iturrate Reyzabal, SM~Hadi Sadati, Hongbin Liu, Sebastien Ourselin, Daniel Leff, Robert~K Katzschmann, Kawal Rhode, and Christos Bergeles.
\newblock Towards a physics-based model for steerable eversion growing robots.
\newblock {\em IEEE Robotics and Automation Letters}, 8(2):1005--1012, 2023.

\end{thebibliography}

\clearpage

\section*{Figures and Tables}
\markright{FIGURES AND TABLES}

\newpage
\begin{table}[h]
    \centering
    \begin{tabular}{r|c|c}
        \textbf{Environment} & \textbf{mean RMSE (mm)} & \textbf{std RMSE (mm)}\\      
        \midrule
        \textbf{Obtuse Angle Contact}      & 22.2 & 5.1 \\
        \textbf{Acute Angle Contact}      & 20.6 & 3.8  \\
        \textbf{Head On Contact}   & 15.8 & 5.2 \\
        \textbf{Physical Validation Solution 1} & 29.8 & 4.1 \\
        \textbf{Physical Validation Solution 2}  & 23.8 & 5.5\\
    \end{tabular}
    \caption{\newtext{Mean and std of RMSE between simulated predictions and deployed robot behavior for single-obstacle and physical validation environments. (Note: $R_\text{vine}$=33.35mm)}}
    \label{tbl:RMSE_results}
\end{table}

\begin{table}[h]
    \centering
    \begin{tabular}{r|ccc|ccc}
        \multicolumn{1}{c}{} & \multicolumn{3}{c}{\textbf{Default}} & \multicolumn{3}{c}{\textbf{No Geo}} \\
        \textbf{Env} & \textbf{Solve} & \textbf{Best} & \textbf{Iter}
                             & \textbf{Solve} & \textbf{Best} & \textbf{Iter} \\[-0.5em]
                    & \textbf{time} & \textbf{cost} & \textbf{time}
                             & \textbf{time} & \textbf{cost} & \textbf{time} \\         
        \midrule
        \textbf{Plus}      & 20.63s & 4.24 & 5.65s  & 19.50s & 4.04 & 5.71s \\
        \textbf{Maze}      & 19.95s & 3.82 & 6.86s  & 23.64s & 3.91 & 6.49s \\
        \textbf{Pickone}   & 15.94s & 3.08 & 6.06s  & 14.66s & 3.16 & 5.88s \\
        \textbf{Needle}    & 46.08s & 5.09 & 4.97s  & 50.50s & 5.38 & 5.00s \\
        \textbf{Long}      & 27.95s & 5.07 & 6.07s  & 27.88s & 5.21 & 5.44s \\
        \textbf{Tube}      & 6.25s  & 1.24 & 5.88s  & 5.73s  & 1.24 & 5.82s \\
    \end{tabular}
    \caption{Comparison of planning performance with (Default) and without (No Geo) geometric heuristic guidance. 
      \textbf{Solve time:} average time to first solution. \textbf{Best cost:} average final cost after optimization. \textbf{Iter time:} average time per iteration of SST*.
      Surprisingly, the geometric heuristic provides minimal performance improvement across evaluated environments. 
      We hypothesize that contact forces the vine robot into ``funnels,'' guiding it along corridors in the environment.}
    \label{tbl:results}
\end{table}

\clearpage
\section*{Supplementary Information}
\markright{SUPPLEMENTARY INFORMATION}
\subsection*{Vine Fabrication}
The vine robot main body and sPAMs are fabricated with LDPE (low-density polyethylene) tubes of 50$\mu$m thickness.
The inflated radius of the vine robot main body $R_\text{vine} = 33.35$mm and the actuator tube radius $R_\text{act} = 17.18$mm. 
We use 3-D printed brackets to constrict the actuator tube to radius $R_c$ at regular intervals $l_0$, forming a chain of sPAMs.
The 3-D printed bracket has a slot through which the actuators were attached to the vine robot tube using double-sided tape adhesive (True Tape LLC., CO, USA).
This method of attachment for sPAMs ensures the actuators do not delaminate from the vine robot tube when pressurized. 
We control the vine robot growth and steering pressure with feedback-controlled pressure regulators (QB3, Proportion Air, IN, USA). 
An Arduino UNO R4 Minima and a custom QB3 shield are used to send pressure commands. 
For motion tracking, green circular markers are attached to the spine of the vine robot and tracked via an overhead camera. 
The individual frames from the collected videos are segmented in Python using image processing tools from OpenCV2.

\subsection*{Tip-steering}\label{tip_steering}
Our actuation method involves attaching sPAM modules along the length of the vine. This restricts our solutions to vines that are practically feasible to fabricate. Furthermore, this requires a different vine for each environment and start/goal pair. A more adaptable method of actuation is tip-steering, where an actuator is only able to affect the tip of the vine robot but moves up as the tip grows.
\newline
\newline
We demonstrate our simulator's ability to model tip-steering in ~\cref{fig:tip_steer}. This is done by stunting vine growth by penalizing changes in vine length. This allows the simulator to apply actuation to the vine while it is not growing, which emulates tip-steering.
\renewcommand\thefigure{S.\arabic{figure}}  

\begin{figure}[H]
    \centering
    \includegraphics[width=0.8\linewidth,trim={0mm 50mm 0mm 50mm}, clip]{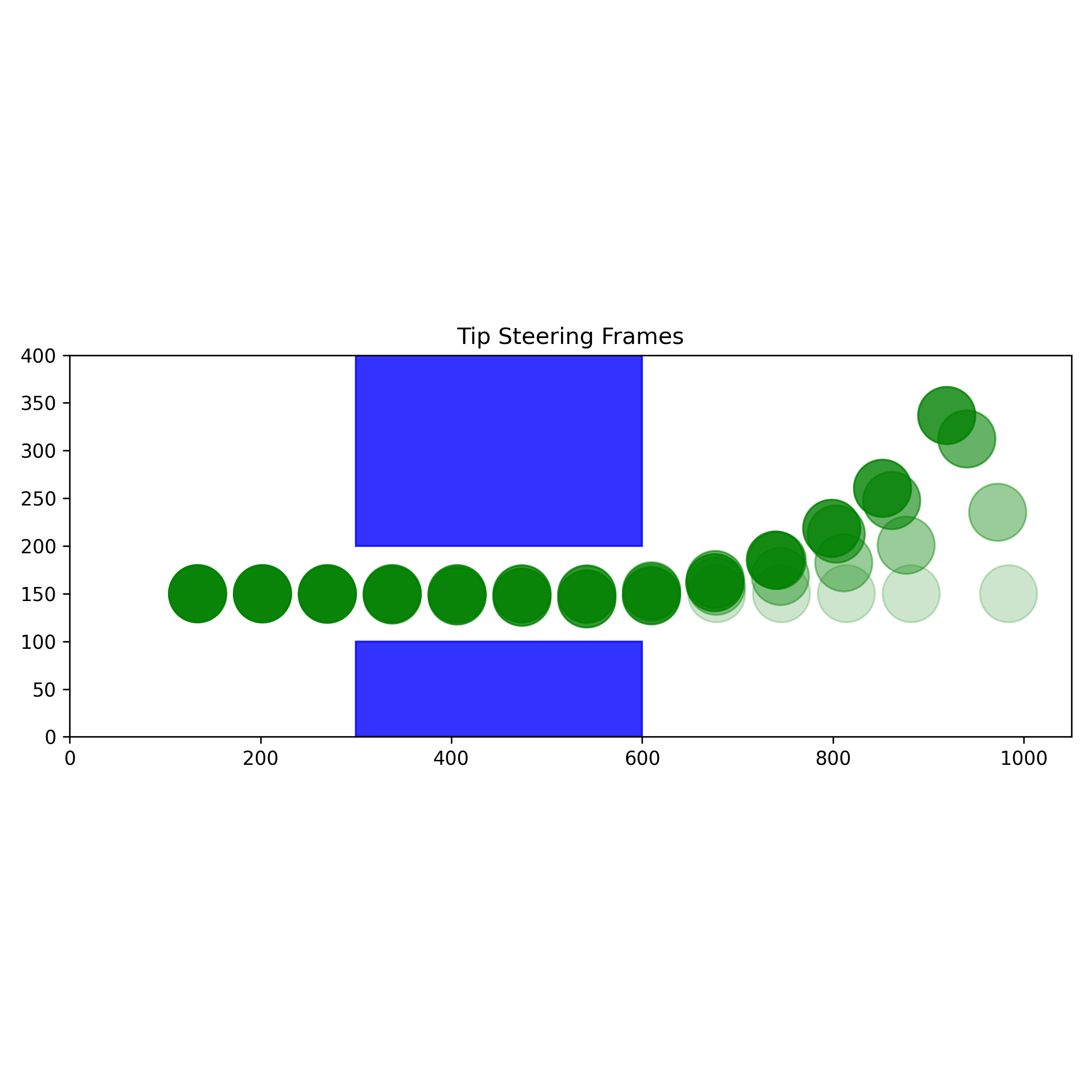}
    \caption{Illustration of tip-steering in simulation. The vine is able to extend through a narrow passage, stop growing, and then apply actuation to the tip. The vine overlay shows the shape of the vine for each timestep that the simulator is applying actuation, but not growth.}
    \label{fig:tip_steer}
\end{figure}

\clearpage

\end{document}